\documentclass[12pt]{report}
\usepackage{biblatex}
\usepackage{graphicx}
\usepackage{amsmath}
\usepackage{amssymb}
\usepackage{algorithm,algpseudocode}
\usepackage{hyperref}
\usepackage{cleveref}
\usepackage{booktabs}
\usepackage{xcolor}
\usepackage{subcaption} 

\usepackage[accsupp]{axessibility}  %

\usepackage{orcidlink}

\usepackage{pifont}%

\newlength{\wid}
\newlength{\mrg}

\newenvironment{Résumé}
    {\clearpage\begin{center}\bfseries Résumé\end{center}}
    {\clearpage}

\newenvironment{Acknowledgement}
    {\clearpage\begin{center}\bfseries Acknowledgement\end{center}}
    {\clearpage}

\begin{document}

\newcommand{\HRule}{\rule{\linewidth}{0.5mm}}

\begin{titlepage}
\thispagestyle{empty}
\begin{center}

\includegraphics[width=0.22\textwidth]{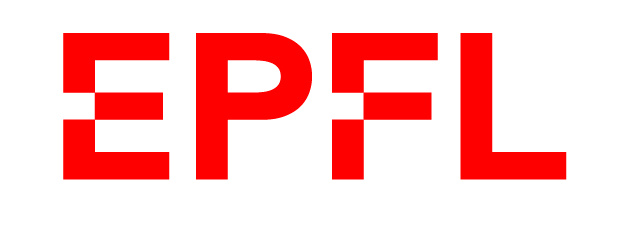}
\includegraphics[width=0.3\textwidth]{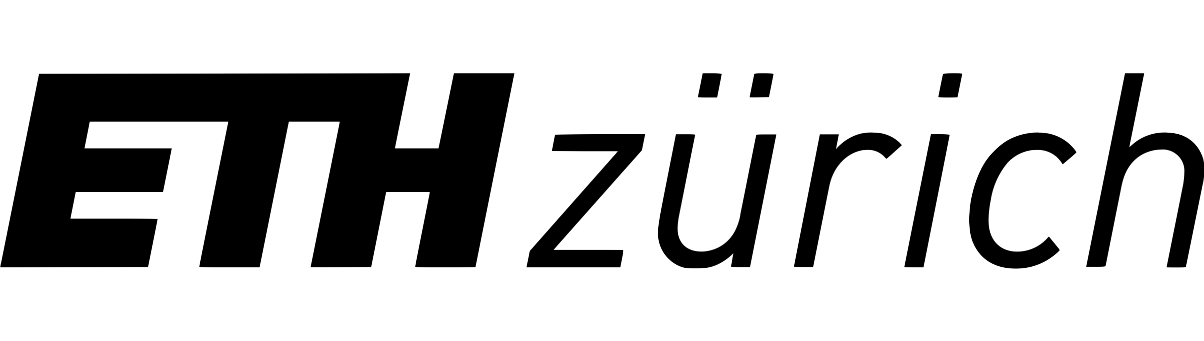}~\\[1cm]
\includegraphics[width=0.12\textwidth]{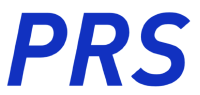}~\\[1cm]

\textsc{\Large Master Thesis}\\[1cm]

\HRule \\[0.4cm]
{ \Huge \bfseries A Modular Framework for Single-View 3D Reconstruction of Indoor Environments \\[0.5cm] }

\HRule \\[1.5cm]

	\begin{minipage}{0.4\textwidth}
		\begin{flushleft}
			\large
			\textit{Author}\\
			Yuxiao Li

            \vspace{2.8cm}
		\end{flushleft}
	\end{minipage}
	~
	\begin{minipage}{0.4\textwidth}
		\begin{flushright}
			\large
			\textit{Advisors}\\
			Prof. Frédéric Kaplan \\
            {\small Digital Humanities Lab}\\
			Prof. Konrad Schindler\\
            {\small Photogrammetry and Remote Sensing Lab}\\
            \vspace{0.5cm}
            \textit{Supervisors}\\
			Dr. Anton Obukhov\\
			Dr. Xi Wang
		\end{flushright}
	\end{minipage}

\vfill
\end{center}
\end{titlepage}

\begin{abstract}
We propose a modular framework for single-view indoor scene 3D reconstruction, where several core modules are powered by diffusion techniques. Traditional approaches for this task often struggle with the complex instance shapes and occlusions inherent in indoor environments. They frequently overshoot by attempting to predict 3D shapes directly from incomplete 2D images, which results in limited reconstruction quality. We aim to overcome this limitation by splitting the process into two steps: first, we employ diffusion-based techniques to predict the complete views of the room background and occluded indoor instances, then transform them into 3D. Our modular framework makes contributions to this field through the following components: an amodal completion module for restoring the full view of occluded instances, an inpainting model specifically trained to predict room layouts, a hybrid depth estimation technique that balances overall geometric accuracy with fine detail expressiveness, and a view-space alignment method that exploits both 2D and 3D cues to ensure precise placement of instances within the scene. This approach effectively reconstructs both foreground instances and the room background from a single image. Extensive experiments on the 3D-Front dataset demonstrate that our method outperforms current state-of-the-art (SOTA) approaches in terms of both visual quality and reconstruction accuracy. The framework holds promising potential for applications in interior design, real estate, and augmented reality. 
\\
\\
Key Words: Single-View Reconstruction, Diffusion Techniques, Amodal Completion, Inpainting Model, Computer Vision
\end{abstract}

\begin{Résumé}
Nous proposons une approche modulaire basée sur des techniques de diffusion pour la reconstruction de scènes d'intérieur en 3 dimensions à partir d'une seule image. Les approches traditionnelles rencontrent souvent des difficultés quand elles font face à des objets aux formes irrégulières ou aux occultations inhérentes aux environnements intérieurs. Celles-ci tentent d'approximer directement la forme des objets à partir d'images incomplètes, conduisant à des reconstructions inexactes. Notre approche vise à surmonter cette limitation en divisant le processus de reconstruction en deux étapes: d'abord, nous exploitons des techniques de diffusion afin de prédire avec précision l'arrière-plan de la pièce et les objets d'intérieur occultés. Ces derniers sont ensuite modélisés en 3 dimensions. Nos contributions sont les suivantes: un module de complétion amodale restaurant complètement les instances occultées, un modèle d'inpainting spécifiquement entraîné pour prédire la disposition des pièces, une technique hybride d'estimation de profondeur équilibrant la géometrie globale de l'environement et les détails de ses éléments, et une méthode d'alignement dans l'espace exploitant à la fois les informations en deux et trois dimensions pour placer avec précision les différents objets de l'environement. Cette approche permet de reconstruire efficacement à la fois les objets en premier plan ainsi que l'arrière-plans de la pièce à partir d'une seule image. Des expériences sur le jeu de données 3D-Front démontrent que notre approche dépasse l'état de l'art en termes de qualité visuelle et de précision de reconstruction. Cette approche semble prometteuse dans les domaines du design d'intérieur, de l'immobilier et de la réalité augmentée.
\\
\\
Mots clés: Reconstruction à partir d'une seule image, Techniques de diffusion, Complétion amodale, Modèle d'inpainting, Vision par ordinateur
\end{Résumé}

\begin{Acknowledgement}
I would like to express my gratitude to those who have supported me throughout my master's thesis:
\begin{itemize}
    \item First of all, I would like to thank Prof. Frédéric Kaplan and Prof. Konrad Schindler for agreeing to host my thesis and provide the computational resources. Their support in ensuring the project's progress and their feedback have been invaluable to my work.
    
    \item Next, I would like to extend my heartfelt gratitude to my supervisors, Dr. Anton Obukhov and Dr. Xi Wang, for their exceptional guidance and support. From our initial brainstorming sessions to their insightful advice on structuring my work, their support has been invaluable. They gave me the freedom to explore different ways to approach the problems while providing the guidance I needed to stay on the right path. I am deeply grateful for their insights and encouragement throughout these months.

    \item I would like to thank Massimiliano Viola for his helpful discussion on Marigold test-time fine-tuning.

    \item Last but not least, I would like to thank my family for their unconditional love and support whenever needed.

\end{itemize}

Zurich, Aug 2024
\end{Acknowledgement}
\tableofcontents
\listoffigures
\listoftables

\chapter{Introduction}
\section{Background and Motivation}
The digitization of instances and environments in 3D has significant potential across industries like interior design and real estate. Traditionally, achieving high-quality 3D reconstruction has relied on photogrammetry~\cite{slama1980manual} or LiDAR~\cite{shan2018topographic}, both of which require specialized equipment and multiple viewpoints. This makes the process costly and less available to non-professionals. Single-view 3D reconstruction, which generates 3D models from a single image, presents a more accessible alternative. However, the recovery of the 3D scene from one single image is inherently difficult due to the lack of multiple perspectives, which limits the ability to correctly estimate depth, handle occlusions, and correct perspective distortions. The model must deal with the unseen part of the instance, leading to great ambiguity in interpreting the 3D object shapes and spatial arrangement. Additionally, single-view methods must heavily rely on indirect cues and learned priors to infer missing information, making them more prone to errors and less reliable in capturing fine details and complex scene layouts compared to multi-view approaches.

Early works in this task primarily represent rooms by combining room layouts with furniture bounding boxes~\cite{du2018learning,  Schwing_2013_ICCV, huang2019perspectivenet}, leading to reconstructions that were relatively basic and offered only contextual information. Later on, methods shift towards leveraging neural networks to infer detailed 3D geometries and their poses from a single image~\cite{dahnert2021panoptic, tulsiani2018factoringshapeposelayout, nie_total3dunderstanding_2020, l_capnet_2018, liu2022towards}, this embodiment enhanced the richness of the reconstructions and expanded their potential applications. However, these methods often struggle with generalization problems, and thus perform poorly on instances or scenes not represented in their training data. They also face challenges with occlusions, where overlapping instances create feature ambiguities and result in errors and distortions in the reconstruction.

Recent advancements in computer vision, particularly those driven by diffusion-based techniques~\cite{ho2020denoising}, have significantly advanced tasks in single-view 3D reconstruction, including single-view human head~\cite{li2024generalizable}, avatar~\cite{zhang2024global, chen2024morphable}, or instances~\cite{xu2024instantmesh, liu2023zero1to3zeroshotimage3d} reconstruction. Furthermore, the introduction of more precise camera calibration~\cite{wang_dust3r_2023} and monocular depth estimation~\cite{wang_dust3r_2023, ke_repurposing_2024} methods that are crucial for single-view 3D reconstruction paves the way for establishing a new framework in this domain. In this work, we focus on the task of \textbf{Single-View Indoor Scene 3D Reconstruction} by establishing a modular framework that leverages these developments in computer vision, it could be decomposed into several parts: instance segmentation and its semantic understanding, monocular depth estimation, camera calibration, amodal completion, background restoration, image-to-3D generation, and view-space alignment. Our method integrates these advancements through a modular architecture, which allows individual components to be replaced easily as more advanced techniques emerge.

The primary contributions of this thesis are as follows:

\begin{itemize}
    \item[--] \textbf{Modular Framework:} We designed a modular framework optimized for single-view indoor scene 3D reconstruction. This framework allows for the easy substitution of modules and ensures consistent updates.
    
    \item[--] \textbf{Amodal Completion Pipeline:} We developed a training-free pipeline for depth-guided amodal completion that specifically addresses the challenges posed by occlusion in indoor scenes.
    
    \item[--] \textbf{Defurnishing Inpainting Model:} We trained a robust inpainting model on synthetic indoor scenes that does not require a perfect inpainting mask. This approach overcomes common issues such as hallucinations and restores the natural look of the reconstructed backgrounds.
    
    \item[--] \textbf{View-Space Alignment:} We introduced a novel view-space alignment technique for scene composition that exploits both 2D and 3D cues to achieve precise object positioning.
    
    \item[--] \textbf{Comprehensive Evaluation:} Through extensive quantitative and qualitative evaluations, we demonstrated that our modular approach delivers generalizable and high-quality full 3D scene reconstructions across various indoor environments.
\end{itemize}

\section{Research Objectives}
This thesis investigates the challenges inherent in single-view 3D reconstruction for indoor scenes. Specifically, it focuses on improving the room layout estimation, reconstructing highly occluded instances, and enhancing view-space alignment for accurate scene composition. The central research questions guiding this thesis are:

\begin{itemize}

\item How can we accurately estimate the layout of an indoor scene and generate a clean room background model, especially when instances significantly obscure parts of the room?

\item How can we reconstruct 3D instances in indoor scenes with significant occlusion without relying on extensive training datasets?

\item What methods can be developed to improve view-space alignment to guarantee that reconstructed elements are positioned correctly within the 3D space relative to the viewer’s perspective?

\item What strategies can be employed to ensure that each component not only performs optimally on its own but also enhances the integration of other components within a modular single-view 3D reconstruction framework?

\end{itemize}

This research aims to develop and evaluate new techniques that address these specific challenges, thereby contributing to more accurate and generalizable single-view indoor scene 3D reconstructions.

\section{Thesis Structure}

This thesis is organized into the following chapters:

\begin{itemize}
    \item \textbf{Chapter 1: Introduction} - Introduces the motivation, background, and objectives of the thesis.
    
    \item \textbf{Chapter 2: Related Work} - Reviews existing literature on single-view indoor scene understanding and 3D scene reconstruction, discussing the strengths and limitations of various approaches.
    
    \item \textbf{Chapter 3: Methodology} - Describes the proposed method in detail, including the method overview, scene understanding, instance reconstruction, and scene composition strategies.
    
    \item \textbf{Chapter 4: Experiments} - Presents the experimental setup, datasets used, and evaluation metrics. It also includes quantitative and Qualitative Comparisons comparing the proposed method with SOTA approaches, along with an ablation study to understand the contribution of each component.
    
    \item \textbf{Chapter 5: Discussion} - Discusses the limitations of the framework and outlines future directions for improvement, along with exploring potential applications.
    
    \item \textbf{Chapter 6: Conclusion} - Summarizes the thesis.
    
\end{itemize}

\chapter{Related Work}

\section{Single-View Indoor Scene Understanding}
Understanding the 3D room geometric information from a single RGB image is the foundation of 3D indoor scene reconstruction. This process involves tasks such as room layout estimation, monocular depth estimation, and 3D object bounding box estimation. These tasks collectively contribute to a comprehensive spatial understanding of the environment by providing precise localization and dimensional information of instances within a scene.

Early approaches in single-view 3D scene understanding primarily focused on solving these tasks independently. For instance, room layout estimation was tackled using standalone methods~\cite{hedau2009recovering, dasgupta2016delay, lee2017roomnet, yan20203d, zheng2020structural, yang2022learning, ibrahem2023st} while instance bounding box estimation followed a similar independent approach~\cite{pan20213d, rukhovich2023tr3d, huang2019perspectivenet, du2018learning, Schwing_2013_ICCV}. Given the inherent connection between room layout and instance positioning, recent efforts have shifted towards joint estimation approaches, where the interdependencies between room layout and instance positioning are exploited to improve accuracy and consistency~\cite{schwing2013box, huang2018cooperative, huang2018holistic, zhang2021holistic}. 

In the domain of monocular depth estimation, initial methods employed multi-scale networks to predict depth from a single image~\cite{eigen_depth_2014}. It has since been improved through techniques such as vision transformers~\cite{yang_depth_2024}. More recently, generative models, particularly diffusion models~\cite{ho2020denoising, song_denoising_2022}, have gained prominence for monocular depth estimation tasks~\cite{ke_repurposing_2024, fu_geowizard_2024}. These models leverage diffusion priors to estimate depth information, demonstrating strong generalization capabilities across a wide range of scenarios.

Our method builds upon these advancements by employing monocular depth estimation for both room layout estimation and instance localization. The process begins with predicting the room background by estimating the pixels occluded by foreground instances with a diffusion-based inpainting model. Subsequently, instance segmentation and the estimated depth are combined to guide the localization of foreground instances, bypassing the need for traditional layout and bounding box estimation methods.

\section{Single-View Indoor Scene 3D Reconstruction}
Building upon the foundations of single-view 3D scene understanding, the task of single-view 3D reconstruction seeks to generate a complete 3D model from a single image. A prominent approach in this domain is panoptic single-view 3D scene reconstruction, which merges scene understanding with scene reconstruction by combining panoptic segmentation and single-view 3D reconstruction~\cite{dahnert2021panoptic, liu_towards_2021, zhang_uni-3d_2023, chu_buol_2024}. This approach directly unifies information obtained from the 2D image, such as semantic and depth information, to regress a detailed 3D scene, thus the correspondence between instances and their spatial locations is always maintained. However, these methods face several limitations. One major challenge is handling complex scenes with significant occlusions or clutter. Additionally, these methods struggle with generalization when encountering in-the-wild scenes that deviate from their training data distribution.

Alternatively, single-view 3D reconstruction could take a compositional approach, where the scene can be parsed into room layout, instance shape, and instance pose. The challenges here are significant mainly due to the inherent ambiguity in interpreting the 3D shapes and poses of instances from a single 2D image, especially when occlusions are frequently present. Early methods use retrieved CAD models of the same category and similar dimensions to represent the original instances in the scenes,~\cite{huang2018holistic, izadinia2017im2cad, mitra2018seethrough}. As this approach only works for instances that closely match the available CAD models in terms of shape, size, and category, it has limited representational power and generalization capability. Later learning-based works regress 3D object shapes from the visible parts of the instances. They employ distinct surface representations, such as voxels~\cite{kulkarni20203drelnetjointobjectrelational, li2019silhouette,  li_geometry_2020, rukhovich_imvoxelnet_2021, tulsiani2018factoringshapeposelayout}, meshes~\cite{huang2018holistic, gkioxari_mesh_2020, gkioxari_learning_2022, kuo_mask2cad_2020, nie_total3dunderstanding_2020}, point clouds~\cite{fan_point_2016, kurenkov_deformnet_2017, l_capnet_2018, mandikal_3d-psrnet_2018}, and implicit surfaces~\cite{chen2024single, liu2022towards}. 
Learning-based methods gained popularity for their ability to reconstruct shapes with high detail. However, because they are trained on specific datasets, they often lack generalization ability to unseen instances shapes. Building on the foundational work of~\cite{poole2022dreamfusion}, which uses generative models for 3D object generation, diffusion-based methods have significantly advanced single-view 3D reconstruction. These methods uses both direct 3D priors~\cite{nichol2022point} and 2D priors~\cite{xu2024instantmesh, liu2023zero1to3zeroshotimage3d, tang2024dreamgaussiangenerativegaussiansplatting, yi2024gaussiandreamerfastgenerationtext, hong2024lrmlargereconstructionmodel, li2024craftsmanhighfidelitymeshgeneration, tochilkin2024triposrfast3dobject, shen_anything-3d_2023} to infer and reconstruct the unseen parts of instances in 3D. A key advantage of diffusion-based methods is their ability to generalize across a wide range of categories without requiring additional training. After obtaining the predicted 3D objects, they are integrated into the scene based on the estimated poses~\cite{huang_holistic_2018, zhang2021holistic}, and could be optionally further fine-tuned~\cite{zhang2021holistic, huang_holistic_2018, chen2019holistic++, izadinia2017im2cad}. 

Our method also takes a compositional strategy. We first reconstruct the background and instances in 3D, then we compose the scene using depth and visual cues to ensure that the reconstructed view corresponds to the original input view.

\chapter{Methodology}
\begin{figure}[h]
    \centering
    \includegraphics[width=1\textwidth]{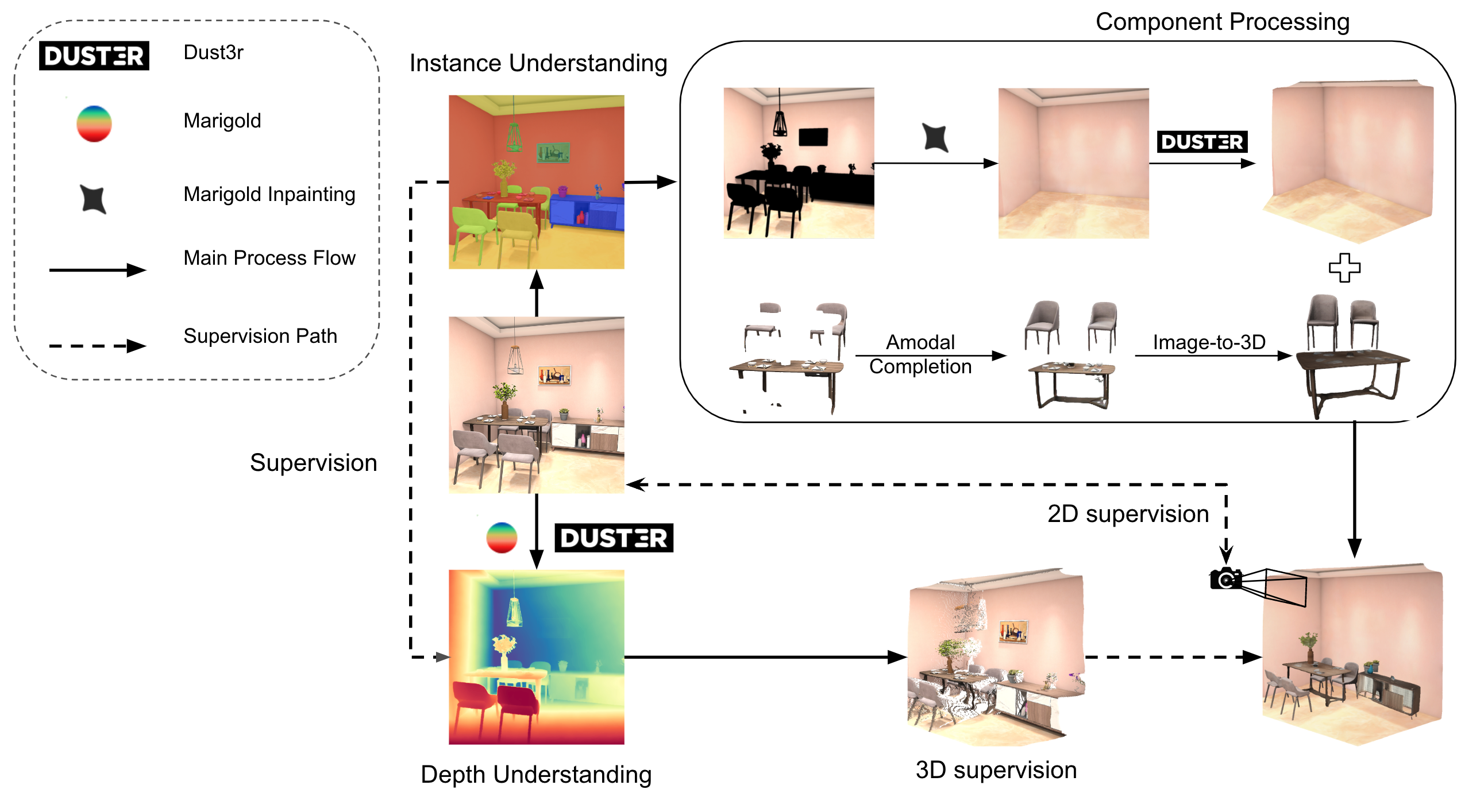}
    \caption[Overview of the proposed pipeline]{\textbf{Overview of the proposed pipeline}. Our method starts with a single image of the indoor scene and, through comprehensive scene understanding, reconstructs the components in 3D. It then integrates both 2D canonical view and 3D points as supervision to compose a complete final reconstruction.}
    \label{fig:overview}
\end{figure}

Our method, as illustrated in~\cref{fig:overview}, reconstructs a holistic scene in 3D from a single RGB image. We structure the method into three interrelated stages: \textbf{Scene Analysis}, \textbf{Component Reconstruction}, and \textbf{Scene Composition}. 

The initial \textbf{Scene Analysis} focuses on extracting critical information for understanding the spatial structure and global context within the scene, such as instance information, camera calibration, and depth information. Building on this foundation, to recover the full geometry of each object, \textbf{Component Reconstruction} employs depth-aware amodal completion to restore occluded regions. The completed 2D instances are then converted into 3D meshes using image-to-3D models. For the room background reconstruction, rather than extrapolating the layout from visible parts, we predict the room layout by "defurnishing" the room image with a custom-trained inpainting model and then lifting it into 3D using depth estimation. The final \textbf{Scene Composition} integrates these reconstructed components into a cohesive 3D scene using both 2D and 3D supervision.



\section{Scene Analysis}
Scene analysis refers to understanding the critical information that guides the component reconstruction, it operates on both the individual instance level and the global scene level. At the instance level, we focus on identifying and segmenting each entity within the scene. At the global level, we analyze global features such as depth and camera parameters. Our method creates a connection between these two levels, where semantic information informs a geometry-aware depth estimation.
\subsection{Instance Understanding}
Instance understanding guides the amodal completion and identifies which parts correspond to foreground instances and which form the background. Traditional semantic segmentation methods~\cite{cheng_masked-attention_2022, jain_oneformer_2022, xie_segformer_2021} produce masks of poor quality. Segment-Anything~\cite{kirillov_segment_2023} offers improved mask quality but struggles to maintain instance cohesion by dividing a single instance into multiple segments. Instance segmentation~\cite{qi_open-world_2022, qi_high-quality_2023} delivers instance mask with high precision, but it lacks the semantic context needed for comprehensive scene understanding.

\begin{figure}[t]
    \centering
    \begin{subfigure}[b]{0.24\textwidth}
        \centering
        \captionsetup{labelformat=empty} 
        \includegraphics[width=\textwidth]{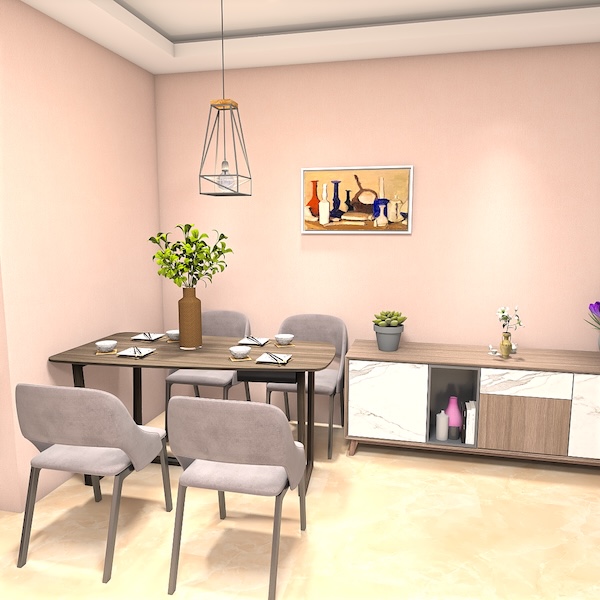}
        \caption{Input Image}
        \label{fig:sub1}
    \end{subfigure}
    \hfill
    \begin{subfigure}[b]{0.24\textwidth}
        \centering
        \captionsetup{labelformat=empty} 
        \includegraphics[width=\textwidth]{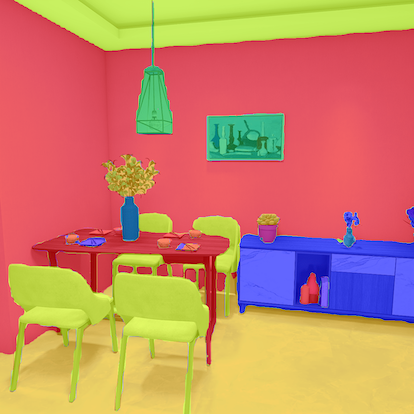}
        \caption{Mask2Former~\cite{cheng_masked-attention_2022}}
        \label{fig:sub2}
    \end{subfigure}
    \hfill
    \begin{subfigure}[b]{0.24\textwidth}
        \centering
        \captionsetup{labelformat=empty} 
        \includegraphics[width=\textwidth]{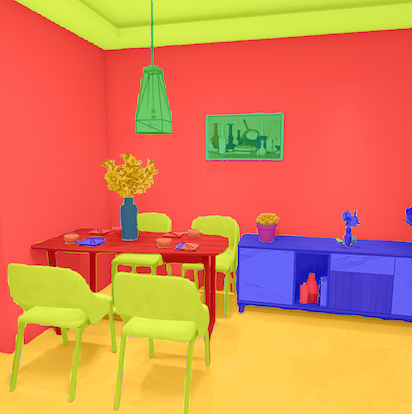}
        \caption{SSA~\cite{chen2023semantic}}
        \label{fig:sub3}
    \end{subfigure}
    \hfill
    \begin{subfigure}[b]{0.24\textwidth}
        \centering
        \captionsetup{labelformat=empty} 
        \includegraphics[width=\textwidth]{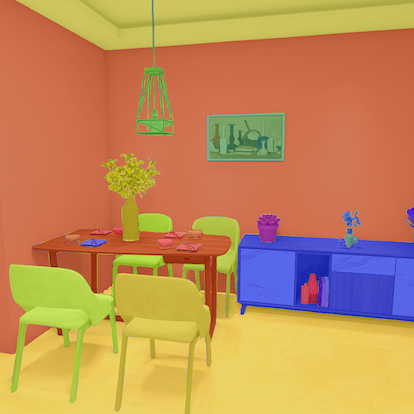}
        \caption{Ours}
        \label{fig:sub4}
    \end{subfigure}
    \caption[Comparison of segmentation methods]{\textbf{Comparison of segmentation methods.} Our method provides high-quality complete masks for each instance while providing semantic information.}
    \label{fig:semantic}
\end{figure}

To address these limitations, we draw inspiration from the Semantic-Segment-Anything (SSA)~\cite{chen2023semantic} framework, aiming to merge the high-quality instance masks generated by CropFormer~\cite{qi_high-quality_2023} with the semantic information from Mask2Former~\cite{cheng_masked-attention_2022}. Our method first employs CropFormer to generate instance masks for the input image $\mathbf{X}$. We then integrate Mask2Former semantic information by filling these instance masks with class labels. Each instance's final label is determined through majority voting within the mask. This strategy allows us to preserve precise instance-level segmentation while guaranteeing that each instance is associated with its semantic category (see~\cref{fig:semantic}). To this end, we obtain the instance mask $\mathbf{O_i}$ together with its semantic label $\mathbf{L_i}$ for each instance in the scene.

\subsection{Depth Understanding and Camera Calibration}
Depth understanding and camera calibration provide crucial information that supports processes such as amodal completion mask refinement, room background reconstruction, and view-space alignment—tasks. These processes demand a high level of depth precision that off-the-shelf monocular depth estimation methods often fail to provide. In our work, we confront the challenge of balancing the overall geometric consistency of depth representation with the need to capture fine structural details. 

As a SOTA method in monocular depth estimation, Marigold~\cite{ke_repurposing_2024} shows superior preservation of detailed structures in its predicted depth map. However, in practical applications, its affine-invariant output requires an estimation of scale and shift. This limitation makes Marigold insufficient for applications requiring exact metric depth measurements. Furthermore, its limited performance in handling distance scene parts leads to the warped geometry of the reconstruction. Dust3r~\cite{wang_dust3r_2023} provides a good estimation of camera calibration and predicts depth maps that correctly capture the overall shapes of indoor scenes. Nonetheless, it frequently produces over-smoothed depth maps, particularly in regions with detailed structural features like chair legs or furniture edges.  While the estimated camera intrinsics \(\mathbf{K}\) can be used as is, the over-smoothed depth maps pose challenges in scenarios where precision in fine-grained details is crucial, such as point cloud registration for view-space alignment, where even minor inaccuracies can result in significant alignment errors.

\subsubsection{Marigold Test-time Fine-Tuning} 
\begin{figure}[t]
    \centering
    \includegraphics[width=1\textwidth]{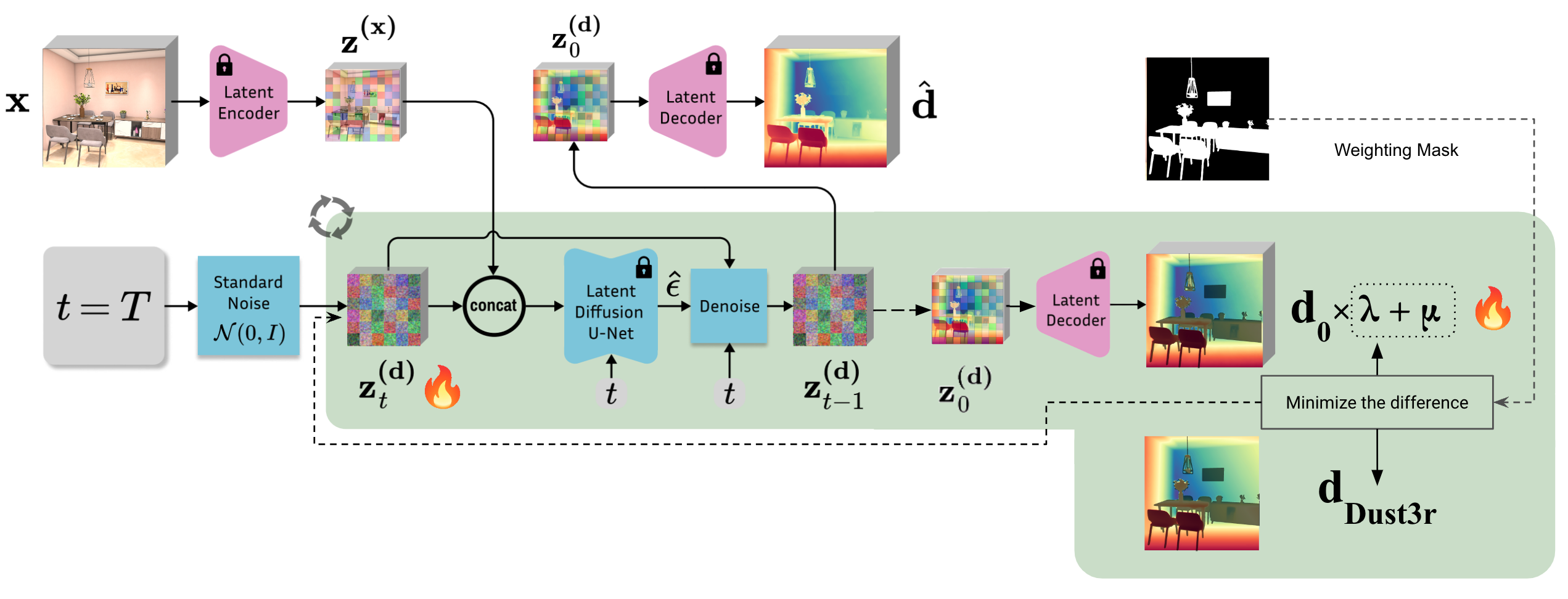}
    \caption[Marigold Test-time Fine-Tuning Schema]{\textbf{Marigold Test-time Fine-Tuning Schema.} This figure is adapted from Original Marigold~\cite{ke_repurposing_2024} Figure 3. Compared to the original inference scheme, we predict the denoised depth latent $\mathbf{z}_{0}^{(\text{d})}$ directly from noise latent $\mathbf{z}_{t}^{(\text{d})}$, followed by further minimizing the difference between the decoded Marigold depth and the Dust3r depth. The gradient is back-propagated to optimize the scale $\lambda$, shift $\mu$, and the noise latent $\mathbf{z}_{t}^{(\text{d})}$, thereby aligning the output Marigold depth with the precise geometric representation provided by the Dust3r depth.}
    \label{fig:marigoldfinetuneprotocal}
\end{figure}

To overcome these limitations, we propose an approach that combines the strengths of Dust3r~\cite{wang_dust3r_2023} and Marigold~\cite{ke_repurposing_2024} shown in~\cref{fig:marigoldfinetuneprotocal}: By fine-tuning the Marigold depth \( D_{\text{Marigold}} \) with the Dust3r depth \( D_{\text{Dust3r}} \), we preserve both the fine details of Marigold depth and the overall geometry correctness of Dust3r depth. Since the Marigold output depth is affine-invariant, we adjust it by estimating the shift $\mu$ as the minimum depth from Dust3r and the scale $\lambda$ as the ratio of Dust3r's maximum to minimum depth values. 
The loss function refines the Marigold output by minimizing the difference between the aligned Marigold depth \( D_{\text{Marigold'}} = \lambda \times D_{\text{Marigold}} + \mu \) and the Dust3r depth \( D_{\text{Dust3r}} \). 
It’s important to note that Dust3r tends to smooth over regions with edges. To prevent Marigold from learning this undesired smoothness in edge areas, we introduce a weighting term in the loss function. This term increases the influence of areas that predominantly consist of flat surfaces like walls, floors, and ceilings, where detail is minimal. As a result, it preserves the overall geometry while maintaining the fine details of furniture and other textured elements. The loss function is thus defined as:

\[
\mathcal{L}(D_{\text{Marigold'}}, D_{\text{Dust3r}}) = \frac{1}{N} \sum_{i=1}^{N} \left( w_i \cdot \left( D_{\text{Marigold'}}(i) - D_{\text{Dust3r}}(i) \right)^2 \right)
\]

where \( w_i \) is a weight factor that increases the influence of the loss in regions with less texture, such as walls, floors, and ceilings, and \( N \) is the total number of pixels in the depth map. By synthesizing these two methods, we overcome the limitations of each and achieve a depth representation that preserves the integrity of delicate structures while maintaining geometric accuracy in the final output.

\section{Reconstruction of Scene Components}
Scene analysis provides the necessary information to reconstruct each individual component in the scene, including the foreground instances and room background. For the foreground instances, we first restore their complete view through amodal completion, then we reconstruct the 3D models with image-to-3D models. For the room background, we follow a similar approach. We train a diffusion inpainting model to first predict the room layout, which is then directly lifted into 3D using predicted depth information and camera intrinsics.
\subsection{Foreground Instance Reconstruction}
In real-world scenarios where multiple pieces of furniture coexist in a living space, instances in a 2D image are often partially occluded by other items. To achieve a complete representation of each instance, we implement amodal completion, a process designed to reconstruct the occluded areas of the instances. Our method utilizes instance segmentation results and depth information to guide a diffusion inpainting model, resulting in the restoration of the full views of the occluded instances.

\begin{figure}[t]
    \centering
    \includegraphics[width=1\textwidth]{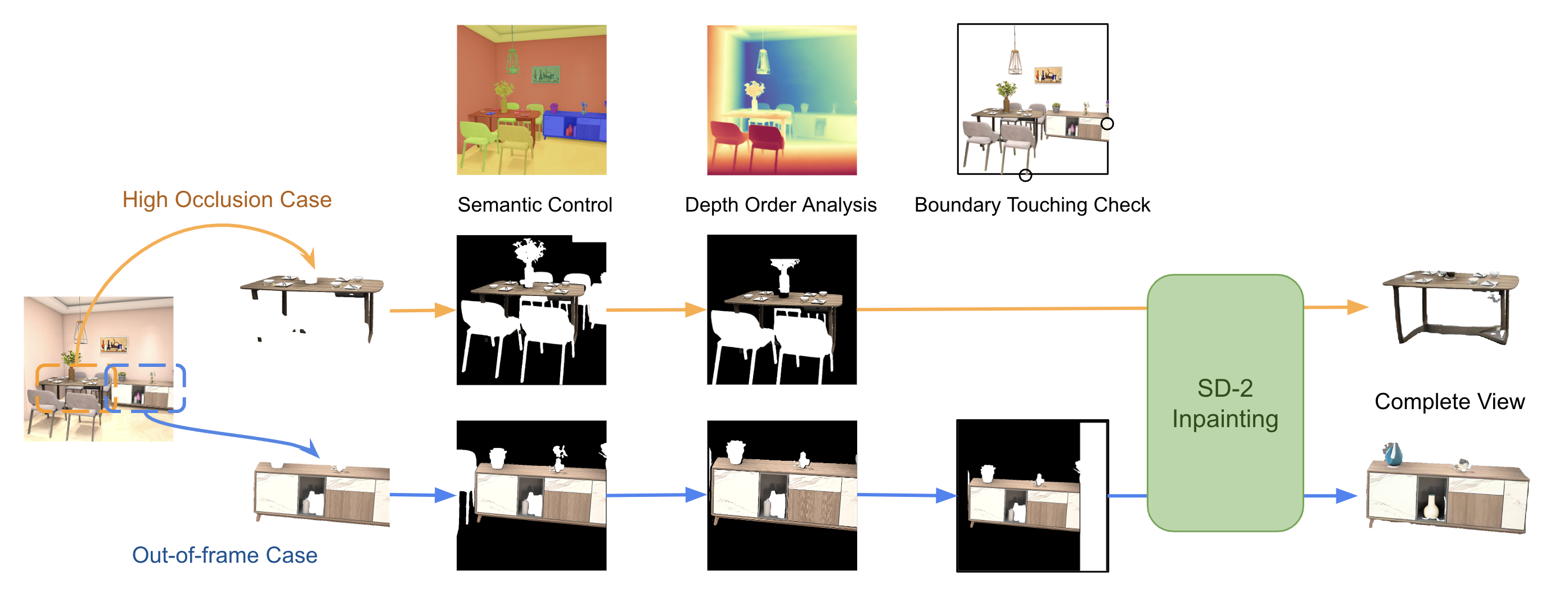}
    \caption[Details of the Amodal Completion module.] {\textbf{Details of the amodal completion module.} We initialize the inpainting mask with the instance masks of neighbouring instances. To refine the inpainting mask, we start by removing pixels that are farther from the camera than the instance being processed. Next, we perform a boundary-touching check, extending the mask to include out-of-frame areas where the instance touches the image boundary. Once the complete inpainting mask is obtained, we use SD2 inpainting to restore the full view of the instance.}
    \label{fig:amodalcompletion}
\end{figure}

As shown in~\cref{fig:amodalcompletion}, first, we initialize the inpainting mask $\mathbf{M_i}$ for the instance $\mathbf{O_i}$ with neighbouring pixels belonging to other instance masks $\mathbf{O_j}$. A key principle in handling occlusion is that an instance can only be occluded by instances that are positioned closer to the camera. To account for this, we perform depth order analysis~\cite{lee_instance-wise_2022}, where we use the depth estimation obtained in the earlier step, denoted as $\mathbf{D}(x, y)$. The inpainting mask $\mathbf{M_i}$ is refined by excluding pixels $(x, y)$ where the depth value $\mathbf{D}(x, y)$ exceeds the maximum depth $D_{\text{max}}^{\mathbf{O_i}}$ of the processed instance. This ensures that only pixels corresponding to instances that are positioned closer to the camera could be retained in the refined mask $\mathbf{M_i'}$.

Additionally, instances could be unoccluded but partially out-of-frame, we identify these cases by checking whether the instance touches the image boundary, denoted as $\mathbf{B}$. When this condition is met, the algorithm extends the inpainting mask $\mathbf{M_i'}$ into the out-of-frame areas. This extension ensures that the entire instance, including portions outside the initial frame, is captured, thereby improving the completeness and accuracy of the inpainting process. The final refined inpainting mask, $\mathbf{M_i''}$, thus integrates depth order analysis with boundary detection to address both occlusion and out-of-frame challenges.
\begin{algorithm}[t]
\caption{Refined Inpainting Mask for Occlusion Handling}
\label{alg:occlusion_handling}
\begin{algorithmic}[1]
\State \textbf{Step 1: Initialize Inpainting Mask}
\State Identify contingent instances $\mathbf{O_j}$ that are adjacent or overlapping with instance $\mathbf{O_i}$.
\State Initialize the inpainting mask $\mathbf{M_i}$ for the instance $\mathbf{O_i}$ by including pixels from these neighbouring masks $\mathbf{O_j}$.

\State \textbf{Step 2: Depth Order Analysis}
\State Perform depth order analysis using depth estimation $\mathbf{D}(x, y)$ to refine the inpainting mask $\mathbf{M_i}$.
\For{each pixel $(x, y)$ in $\mathbf{M_i}$}
    \If{$\mathbf{D}(x, y) > D_{\text{max}}^{\mathbf{O_i}}$}
        \State Exclude pixel $(x, y)$ from $\mathbf{M_i'}$ 
    \EndIf
\EndFor

\State \textbf{Step 3: Boundary Detection and Extension}
\For{each instance $\mathbf{O_i}$ in the image}
    \If{instance $\mathbf{O_i}$ touches the image boundary $\mathbf{B}$}
        \State Extend the inpainting mask $\mathbf{M_i'}$ to cover out-of-frame areas
    \EndIf
\EndFor

\State \textbf{Step 4: Final Mask Refinement}
\State Combine depth order refinement and boundary detection to produce the final refined inpainting mask $\mathbf{M_i''}$.

\State \textbf{return} Refined inpainting mask $\mathbf{M_i''}$
\end{algorithmic}
\end{algorithm}

After we obtain the inpainting mask, we use it to guide the amodal completion with diffusion inpainting model—specifically, Stable-Diffusion-2 (SD2) Inpainting~\cite{Rombach_2022_CVPR}. To avoid generating unwanted artefacts, we insert the processed instance  $\mathbf{O_i}$ into a background of random noise. The resulting noised image $\mathbf{N_i}$ along with the refined inpainting mask $\mathbf{M_i''}$ is then fed into the SD2 Inpainting model. The inpainting process is guided by a prompt informed by the semantic label $\mathbf{L_i}$ of the processed instance to generate the complete view $\mathbf{C_i}$ of the processed instance $\mathbf{O_i}$. The complete view $\mathbf{C_i}$ could be transformed to 3D using image-to-3D models. For the image-to-3D generation, we choose InstantMesh~\cite{xu2024instantmesh} due to its quick processing time and strong generalization capabilities. At this stage, we have obtained 3D mesh models \(\mathcal{M}_i\) that represent the complete instances within the view.

\subsection{Room Background Reconstruction}

Room background refers to the structural elements of a room, such as walls, ceilings, and floors. While these components generally have simple geometry, estimating the room background by extrapolating from visible parts or using~\cite{liu2022towards, dogaru_generalizable_2024} a basic cuboid representation~\cite{nie_total3dunderstanding_2020, zhang2021holistic} usually yield unsatisfactory results. We follow a similar approach as in "Foreground Instance Reconstruction" by predicting the room background through "defurnishing" the space using inpainting techniques, and then converting the predicted room background into 3D. 

The defurnishing inpainting process involves placing an inpainting mask over the foreground instances and using prompts such as "unfurnished empty room" to defurnish the room. Nevertheless, off-the-shelf inpainting models are typically trained on large datasets of images containing fully furnished indoor scenes, making them more adept at modifying instances and completing scenes rather than producing empty spaces. Consequently, when these models are intended for defurnishing, they often introduce unwanted artefacts such as hallucinations~\cite{slavcheva_empty_2024}. Stable Diffusion Inpainting~\cite{podell2023sdxlimprovinglatentdiffusion, noauthor_diffusersstable-diffusion-xl-10-inpainting-01_nodate, Rombach_2022_CVPR}, for example, may attempt to "explain" the leftover artefacts, like shadows or remnants of instances left out by an imperfect inpainting mask, resulting in the creation of unwanted elements in the defurnished room. To address this issue, the inpainting mask could be expanded to cover areas with shadows or other residual elements. Nonetheless, this approach risks inadvertently obscuring important room structure features that are essential for maintaining the original room layout during the inpainting process, such as room corners or skirting lines. Overextending the mask could result in a room layout that deviates significantly from the original, thereby compromising the accuracy of the defurnished room.

Given the challenges of using Stable Diffusion Inpainting to generate defurnished rooms, we decide to develop our defurnishing inpainting model. To this end, we repurpose the original Marigold~\cite{ke_repurposing_2024} for this task: instead of predicting single-channel depth from an RGB image, we adapt Marigold to accept an RGBA image as input, where the RGB channel corresponds to input image \( \mathbf{X} \) and alpha channel represents the inpainting mask $\sum_{i=1}^{n} \mathbf{O_i}$, and produces an RGB image of the defurnished room \( \mathbf{Y} \). After generating the defurnished room image \( \mathbf{Y} \), we directly lift it into 3D using Dust3r~\cite{wang_dust3r_2023} without Marigold fine-tuning, as empty room backgrounds typically have simple geometric structures. Until now, we have obtained the 3D mesh model \(\mathcal{M}_\text{background}\) for the room background.

\begin{figure}[t]
    \centering
    \includegraphics[width=1\textwidth]{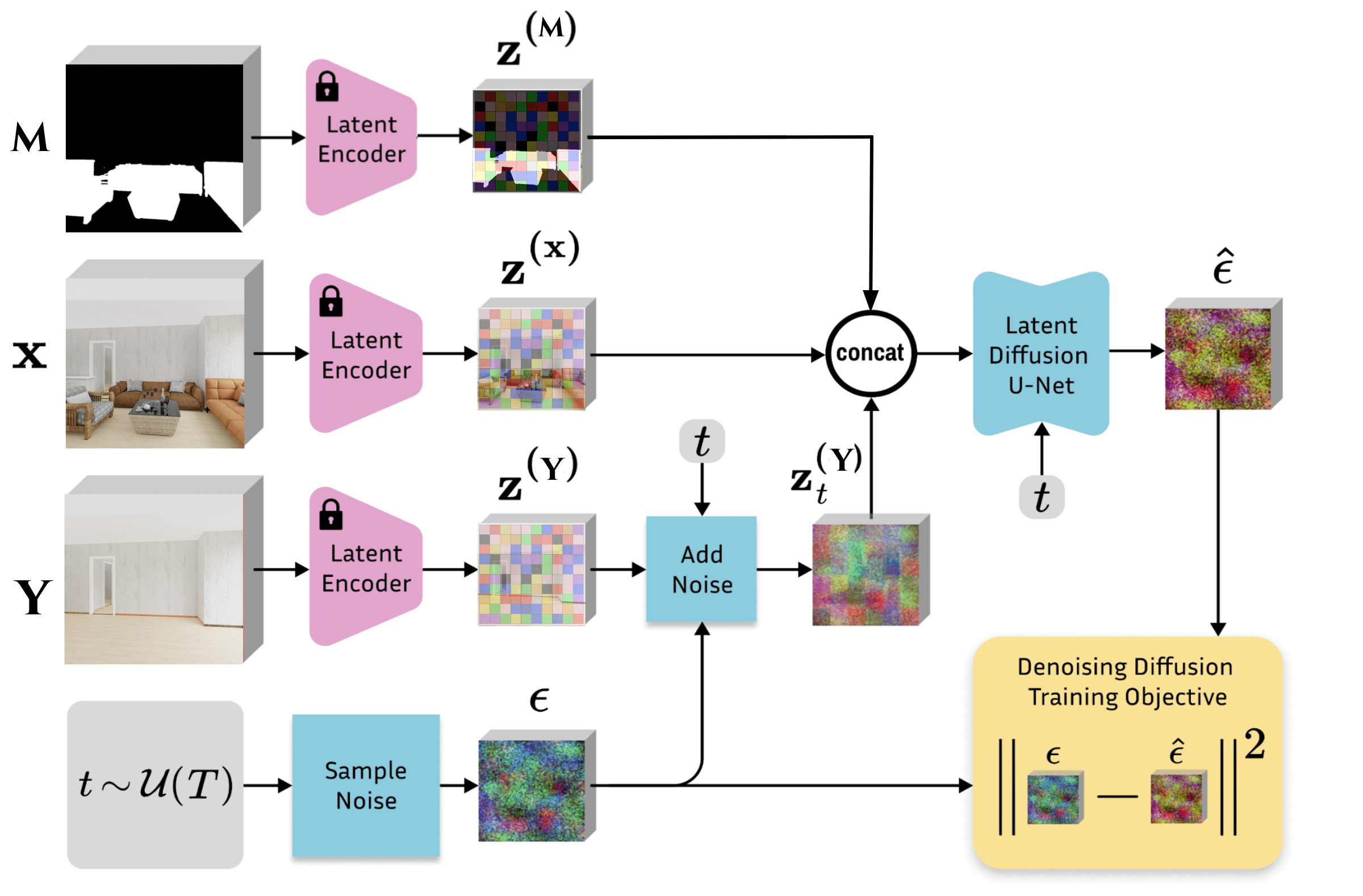}
    \caption[Marigold inpainting fine-tuning protocol]{\textbf{Overview of the Marigold inpainting fine-tuning protocol.} This figure is adapted from Original Marigold~\cite{ke_repurposing_2024} Figure 2. Compared with the original Marigold, we further include the inpainting mask \( \mathbf{M} \), and instead of predicting the depth \( \mathbf{d} \), we aim to predict the empty room \( \mathbf{Y} \).}
    \label{fig:marigoldinpaintingoverview}
\end{figure}

\subsubsection{Training Data Preparation}
To generate the training data focusing on indoor environments, we utilize BlenderProc~\cite{Denninger2023} to render synthetic scenes from the 3D-FRONT Dataset~\cite{fu20213d}, a synthetic dataset focused on indoor scenes. We generate image pairs with and without foreground furniture by first rendering the complete, fully furnished scene. Next, we remove all visible foreground furniture instances and re-rendered the scene as an empty room, maintaining the same camera parameters. This method ensured consistent perspective and layout between the furnished and empty scenes.

The camera is sampled within a height range of 1.4 to 1.9 meters, corresponding to typical human eye level, to align the rendered scenes with a human perspective. We also implement proximity checks to maintain a minimum camera distance of 3 meters from any instance to ensure that the scene is adequately populated and reflects common viewing perspectives. To increase the realism of the training data, we introduce varying lighting conditions across different renders to simulate a wide range of real-world scenarios. In addition to RGB images, we also render the corresponding instance segmentation maps to obtain training ground truth masks for inpainting. In practical applications, we often rely on standard image segmentation algorithms to generate inpainting masks, which can be prone to quality issues like inaccurate boundaries or incomplete coverage. To mitigate these limitations, we manipulate our inpainting masks by randomly enlarging and shrinking them to ensure that our model becomes more robust to imperfections in the inpainting masks, allowing it to perform well even when the segmentation masks are imperfect.

This comprehensive setup results in 17315 image pairs at a resolution of 1024x1024 pixels, providing a photorealistic dataset for training and evaluating our defurnishing inpainting model.

\subsubsection{Network Architecture}
We adapt the network architecture from the original Marigold. As shown in~\cref{fig:marigoldinpaintingoverview}, the encoder in Marigold is designed for 3-channel images, we replicate the single-channel alpha inpainting mask into a three-channel format. The encoded mask image $\mathbf{z}^{(\mathrm{M})}$ is then concatenated with the encoded room image $\mathbf{z}^{(\mathrm{X})}$ and encoded empty room image with noise $\mathbf{z}_{t}^{(\mathrm{Y})}$. This combined latent representation is used throughout the training process, allowing the model to integrate transparency data. The rest of the setup follows the standard Marigold pipeline.

\section{Scene Composition}
The room background is reconstructed in camera space using the estimated camera intrinsics \(\mathbf{K}\) and depth information \( D_{\text{empty}} \), while the foreground instances are reconstructed in a canonical space based on the Zero 1-2-3 camera settings~\cite{liu2023zero1to3zeroshotimage3d}. This approach introduces scale differences between the background and foreground elements. To correctly position the instance within the reconstructed room background, we estimate the scale factor \(s_i\) by using reference points \(\ P_{i}^{\text{view}} \) derived from the depth data \(D(\mathbf{u})\), camera intrinsics \(\mathbf{K}\) and instance mask $\mathbf{O_i}$, which provide initial guidance on the furniture's position and scale. After this initial alignment, we aim to obtain the transformation \(\mathcal{T}(R, t)\) that registers the visible points of the furniture meshes $\mathcal{M}_{i}^{\text{view}}$ with the corresponding points from the initial view \(\ P_{i}^{\text{view}} \) through ICP algorithm. This registration is accomplished by extracting visible points through ray-mesh intersections based on the current camera viewpoint. The resulting transformation \(\mathcal{T}(R, t)\) is then applied to the full mesh \(\mathcal{M}_i\).

To further refine the scene composition, we employ a differentiable renderer~\cite{ravi2020pytorch3d} to optimize the scale, orientation, and position of each instance, using the 2D canonical input view for supervision to enforce that the rendered instance aligns closely with its input view. The process begins by generating an initial rendering of the foreground instance mesh based on the previous alignment steps. This rendered instance is then compared to its counterpart in the original 2D image. By minimizing MSE Loss, which measures the pixel-level difference between the rendered and original views, we iteratively fine-tune the instance's parameters until the rendered output faithfully reproduces the original. The loss function is defined as

\[
\mathcal{L}(t_i, s_i, R_i) = \mathbb{E}\left[\left\|\mathbf{X}_{\text{rendered}}(t_i, s_i, R_i, \mathbf{K}) - \mathbf{X}_{\text{O}_i}\right\|_2^2\right]
\]

where \( t_i = (x_i, y_i, z_i) \) is the translation vector for the \(i\)-th instance, \( s_i \) denotes the scaling factor, and \( R_i \) is the rotation matrix. The term \(\mathbf{X}_{\text{rendered}}(t_i, s_i, R_i, \mathbf{K})\) refers to the rendered image obtained from the 3D mesh using the pose parameters \((t_i, s_i, R_i)\) and camera intrinsics \(\mathbf{K}\). \(\mathbf{X}_{\text{O}_i}\) is the canonical instance view from the 2D image, and \(\left\| \cdot \right\|^2\) denotes the L2 loss, which measures the pixel-wise difference between the rendered and original images.

\chapter{Experiments}

\section{Marigold Inpainting Model}
\subsection{Implementation}
We adopt the implementation details from the original Marigold~\cite{ke_repurposing_2024} to train our defurnishing model. The training is conducted over 20,000 iterations with a batch size of 32. To accommodate a single GPU environment, gradient accumulation is performed across 16 steps. The entire training process requires approximately 3 days to converge on an Nvidia RTX 4090 GPU. For inference, we employ the DDPM scheduler~\cite{ho2020denoising} with 50 sampling steps, consistent with the original Marigold setup\cite{ke_repurposing_2024}. Our final predictions do not require any ensemble methods.

\subsection{Evaluation}

We assess the quality of our inpainting method using both quantitative and qualitative evaluations. Quantitatively, we conduct experiments on 1,000 high-quality indoor scene images rendered from the 3D-Front dataset~\cite{fu20213d}, each with a resolution of 1200 x 1200 pixels. Qualitatively, we examine various in-the-wild examples to evaluate the generalization capability of our method. We compare our results with other SOTA inpainting methods, including SD2 Inpainting~\cite{Rombach_2022_CVPR}, SDXL Inpainting~\cite{podell2023sdxlimprovinglatentdiffusion, noauthor_diffusersstable-diffusion-xl-10-inpainting-01_nodate}, and LaMa~\cite{suvorov2021resolutionrobustlargemaskinpainting}. 

\begin{itemize}
    \item \textbf{SD2 Inpainting~\cite{Rombach_2022_CVPR}}: Builds on stable-diffusion-2-base, further trained with enhanced mask generation techniques and latent VAE conditioning for improved inpainting performance.
    \item \textbf{SDXL Inpainting~\cite{podell2023sdxlimprovinglatentdiffusion, noauthor_diffusersstable-diffusion-xl-10-inpainting-01_nodate}}: Extends stable-diffusion-xl-base-1.0 by incorporating additional channels and high-resolution training for inpainting tasks.
    \item \textbf{LaMa~\cite{suvorov2021resolutionrobustlargemaskinpainting}}: Tackles large missing areas and complex structures using fast Fourier convolutions (FFCs) for an image-wide receptive field, combined with high receptive field perceptual loss and large training masks for improved inpainting performance.
\end{itemize}

It's important to note that the inpainting masks are obtained using the semantic segmentation method introduced earlier, which highlights that our method performs well even without perfect inpainting masks.
\subsubsection{Quantitative Evaluation}

The quantitative evaluation is conducted using four key metrics: Peak Signal-to-Noise Ratio (PSNR), Mean Squared Error (MSE) for measuring absolute differences, and Structural Similarity Index Measure (SSIM)~\cite{wang_image_2004} along with Learned Perceptual Image Patch Similarity (LPIPS)~\cite{zhang_unreasonable_2018} for evaluating perceptual differences. PSNR and MSE are traditional image quality metrics that measure the absolute difference between the image pairs, where higher PSNR and lower MSE values indicate better performance. SSIM assesses the structural similarity between the image pairs, with higher values reflecting better preservation of the image's structural integrity. LPIPS, a more recent metric, measures perceptual similarity, capturing human-perceived differences between images; lower LPIPS values indicate higher perceptual fidelity. The results are shown in table~\ref{tab:inpainting}. Given that our method is the only one that removes foreground furniture without introducing artefacts, it demonstrates a clear advantage over other approaches in metrics that assess absolute differences, including PSNR and MSE. SD2 and SDXL inpainting methods often suffer from hallucination issues, which significantly degrade the overall inpainting quality across all metrics. While LaMa typically inpaints masked areas by replicating textures from adjacent regions, resulting in strong structural consistency and achieving the highest SSIM scores, its performance declines on LPIPS that better reflects human-perceived similarity.

\begin{table}[ht]
\centering
\begin{tabular}{lcccccc}
\toprule
\textbf{Method} & \textbf{PSNR $\uparrow$} & \textbf{MSE $\downarrow$} & \textbf{SSIM $\uparrow$} & \textbf{LPIPS $\downarrow$} \\
\midrule
LaMa~\cite{suvorov2021resolutionrobustlargemaskinpainting} & 22.41 & 0.080 & {\textbf{0.917}} & 0.097 \\
SD2 inpainting~\cite{Rombach_2022_CVPR} & 23.10 & 0.074 & 0.847 & 0.120 \\

SDXL inpainting~\cite{podell2023sdxlimprovinglatentdiffusion, noauthor_diffusersstable-diffusion-xl-10-inpainting-01_nodate} & 22.40 & 0.083 & 0.869 & 0.114  \\

\midrule
{Ours} & {\textbf{28.35}} & {\textbf{0.041}} & 0.879 & {\textbf{0.073}} \\
\bottomrule
\end{tabular}
\caption[Quantitative Evaluation of inpainting methods for Empty Room Prediction]{\textbf{Quantitative Evaluation of Inpainting Methods for Empty Room Prediction.} Our method demonstrates a significant advantage over other approaches in PSNR and MSE, which assess absolute pixel differences. Additionally, it achieves the highest performance in LPIPS, reflecting superior human-perceived similarity.}
\label{tab:inpainting}
\end{table}

\subsubsection{Qualitative Evaluation}

To assess the visual quality and generalization capability of our model, we evaluate it on images that are not included in the training data, using examples from the 3D-Future~\cite{fu20213dfuture}, Hypersim~\cite{roberts:2021}, and real-world images taken with mobile phones. This evaluation demonstrates the robustness of our model across diverse environments, highlighting its capability to generate clean, unblemished empty rooms and handle various scene types.

\begin{figure*}
    \centering    
    \setlength{\wid}{0.195\textwidth}
    \setlength{\mrg}{-0.2cm}
    \begin{tabular}{@{}ccccc@{}}
        \includegraphics[width=\wid]{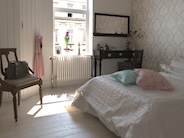} &
        \hspace{-3mm}\includegraphics[width=\wid]{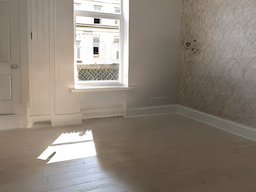} &
        \hspace{-3mm}\includegraphics[width=\wid]{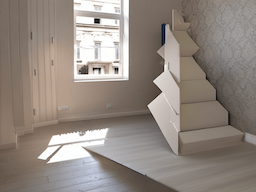} &
        \hspace{-3mm}\includegraphics[width=\wid]{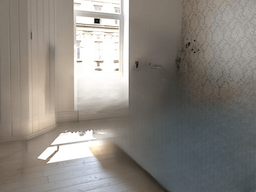} &
        \hspace{-3mm}\includegraphics[width=\wid]{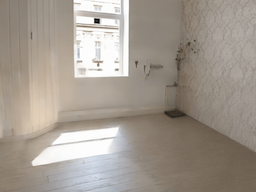} 
        \\ 
        \includegraphics[width=\wid]{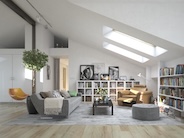} &
        \hspace{-3mm}\includegraphics[width=\wid]{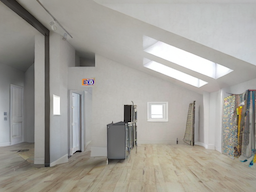} &
        \hspace{-3mm}\includegraphics[width=\wid]{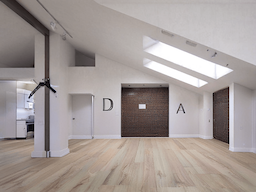} &
        \hspace{-3mm}\includegraphics[width=\wid]{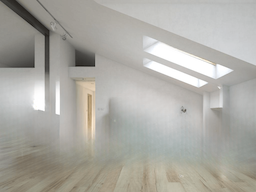} &
        \hspace{-3mm}\includegraphics[width=\wid]{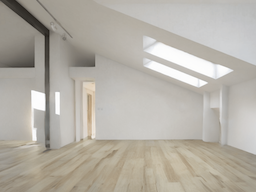} 
        \\
        \includegraphics[width=\wid]{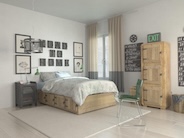} &
        \hspace{-3mm}\includegraphics[width=\wid]{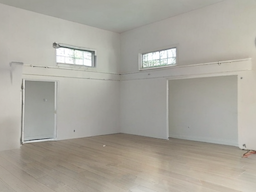} &
        \hspace{-3mm}\includegraphics[width=\wid]{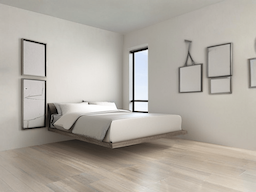} &
        \hspace{-3mm}\includegraphics[width=\wid]{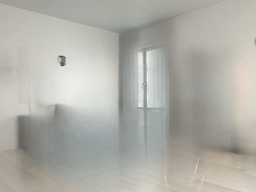} &
        \hspace{-3mm}\includegraphics[width=\wid]{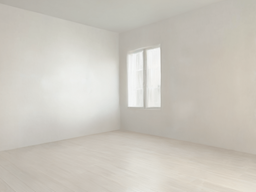} 
        \\
        Input Image & 
        SD2~\cite{Rombach_2022_CVPR}& 
        SDXL ~\cite{podell2023sdxlimprovinglatentdiffusion, noauthor_diffusersstable-diffusion-xl-10-inpainting-01_nodate}& 
        LaMa~\cite{suvorov2021resolutionrobustlargemaskinpainting}& 
        Ours
    \end{tabular}
    \caption[Qualitative Comparison of inpainting methods for Hypersim images~\cite{roberts:2021}]{\textbf{Qualitative Comparison of inpainting methods} for Hypersim images~\cite{roberts:2021}}
    \label{fig:inpainting_hypersim}
\end{figure*}

\begin{figure*}
    \centering    
    \setlength{\wid}{0.195\textwidth}
    \setlength{\mrg}{-0.2cm}
    \begin{tabular}{@{}ccccc@{}}
        \includegraphics[width=\wid]{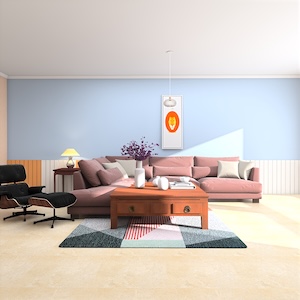} &
        \hspace{-3mm}\includegraphics[width=\wid]{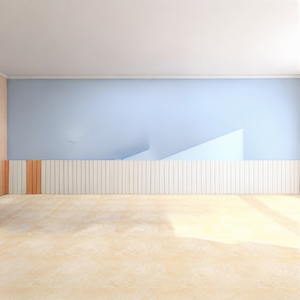} &
        \hspace{-3mm}\includegraphics[width=\wid]{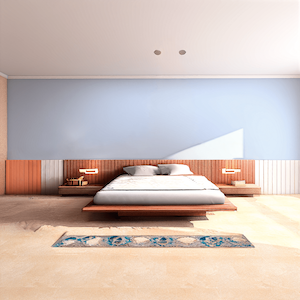} &
        \hspace{-3mm}\includegraphics[width=\wid]{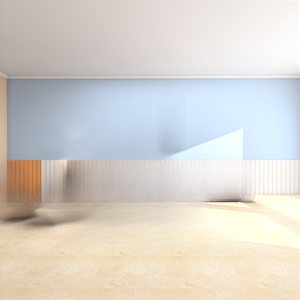} &
        \hspace{-3mm}\includegraphics[width=\wid]{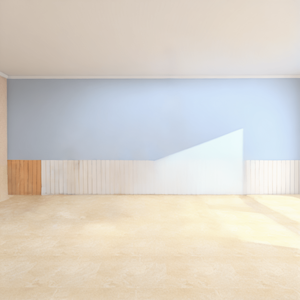} 
        \\ 
        \includegraphics[width=\wid]{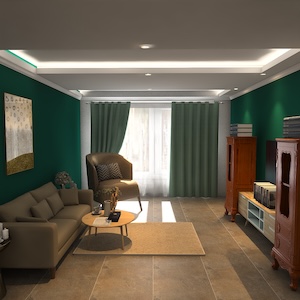} &
        \hspace{-3mm}\includegraphics[width=\wid]{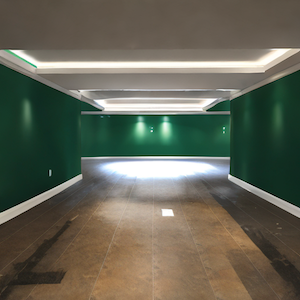} &
        \hspace{-3mm}\includegraphics[width=\wid]{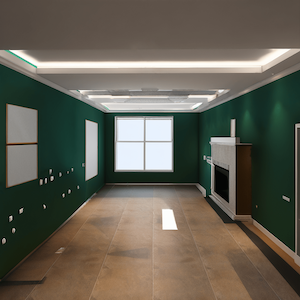} &
        \hspace{-3mm}\includegraphics[width=\wid]{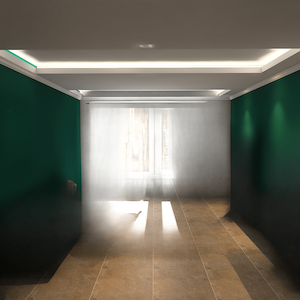} &
        \hspace{-3mm}\includegraphics[width=\wid]{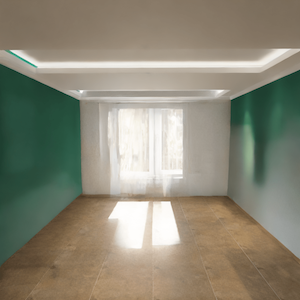} 
        \\
        \includegraphics[width=\wid]{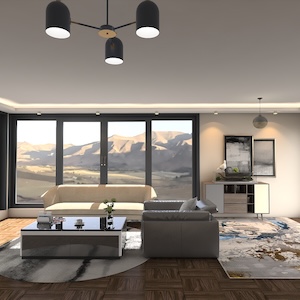} &
        \hspace{-3mm}\includegraphics[width=\wid]{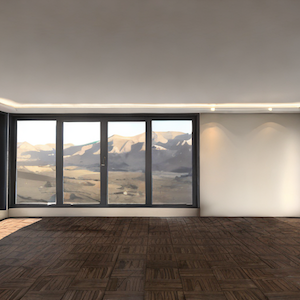} &
        \hspace{-3mm}\includegraphics[width=\wid]{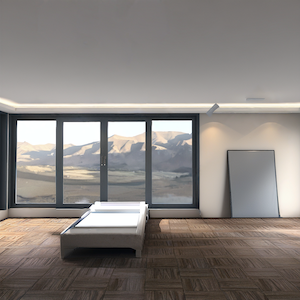} &
        \hspace{-3mm}\includegraphics[width=\wid]{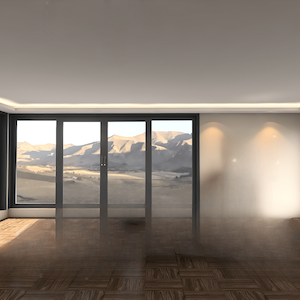} &
        \hspace{-3mm}\includegraphics[width=\wid]{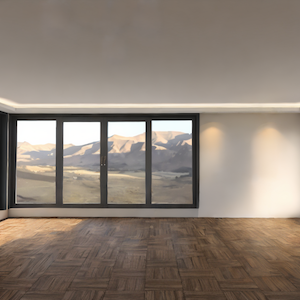} 
        \\
        Input Image & 
        SD2~\cite{Rombach_2022_CVPR}& 
        SDXL~\cite{podell2023sdxlimprovinglatentdiffusion, noauthor_diffusersstable-diffusion-xl-10-inpainting-01_nodate} & 
        LaMa~\cite{suvorov2021resolutionrobustlargemaskinpainting}& 
        Ours
    \end{tabular}
    \caption[Qualitative Comparison of inpainting methods for 3D-Future images~\cite{fu20213dfuture}]{\textbf{Qualitative Comparison of inpainting methods} for 3D-Future images~\cite{fu20213dfuture}}
    \label{fig:inpainting_3dfutre}
\end{figure*}
\begin{figure*}
    \centering    
    \setlength{\wid}{0.195\textwidth}
    \setlength{\mrg}{0cm}
    \begin{tabular}{@{}ccccc@{}}
        \includegraphics[width=\wid]{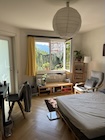} &
        \hspace{-3mm}\includegraphics[width=\wid]{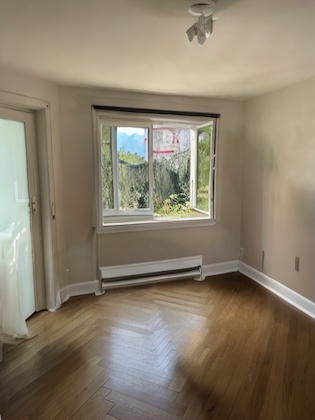} &
        \hspace{-3mm}\includegraphics[width=\wid]{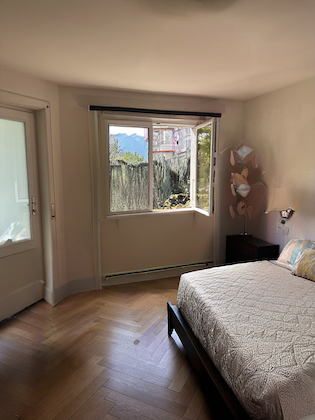} &
        \hspace{-3mm}\includegraphics[width=\wid]{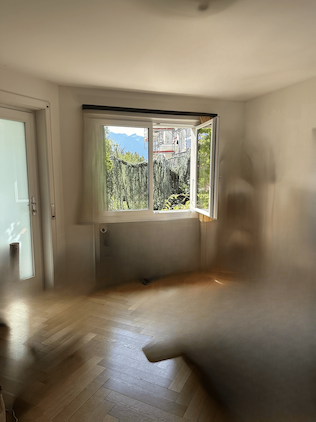} &
        \hspace{-3mm}\includegraphics[width=\wid]{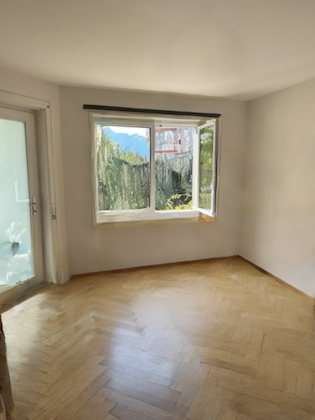} 
        \\ 
        \includegraphics[width=\wid]{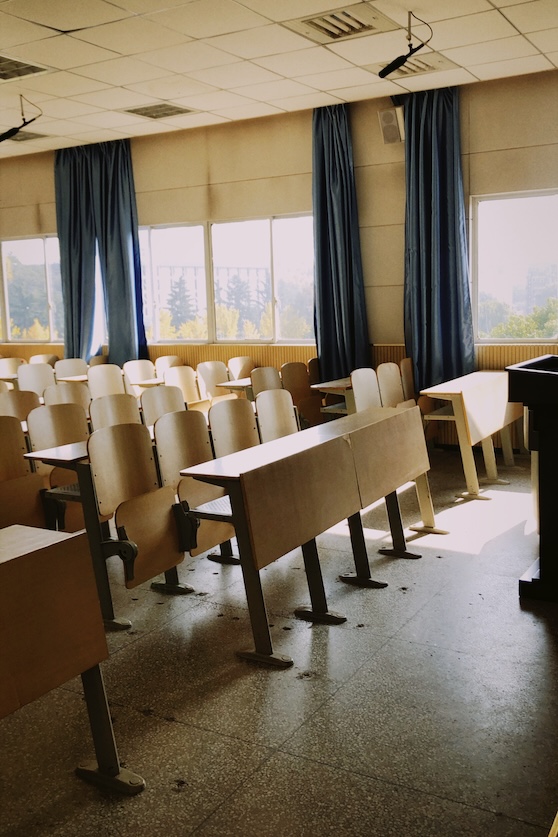} &
        \hspace{-3mm}\includegraphics[width=\wid]{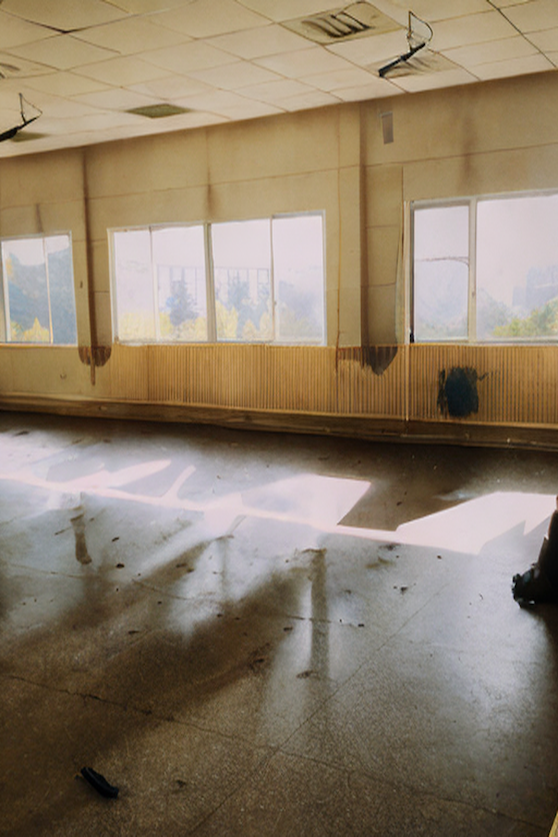} &
        \hspace{-3mm}\includegraphics[width=\wid]{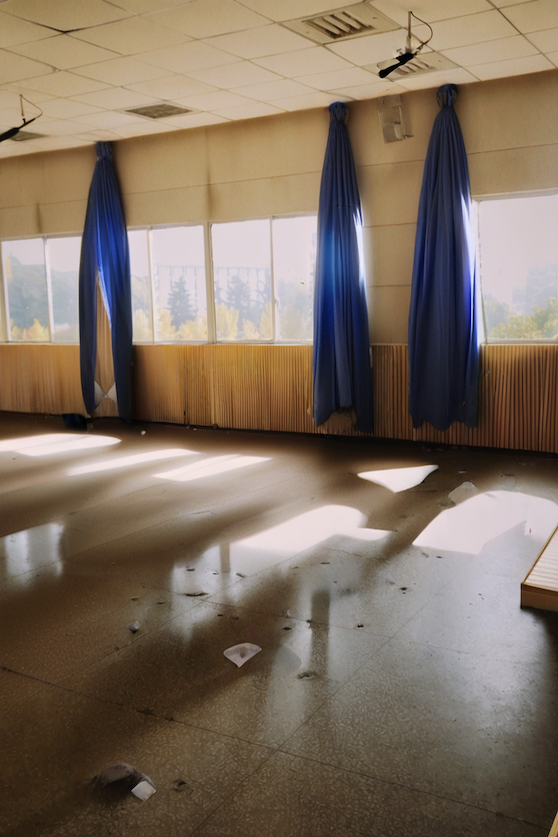} &
        \hspace{-3mm}\includegraphics[width=\wid]{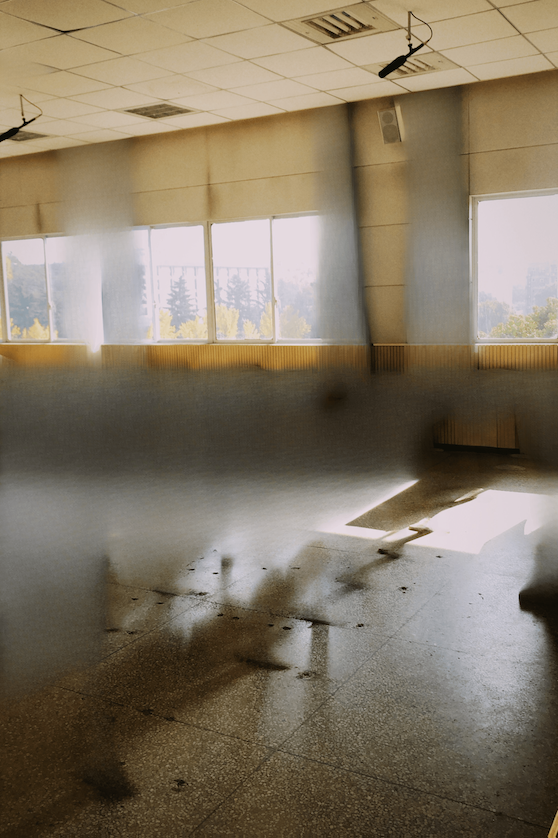} &
        \hspace{-3mm}\includegraphics[width=\wid]{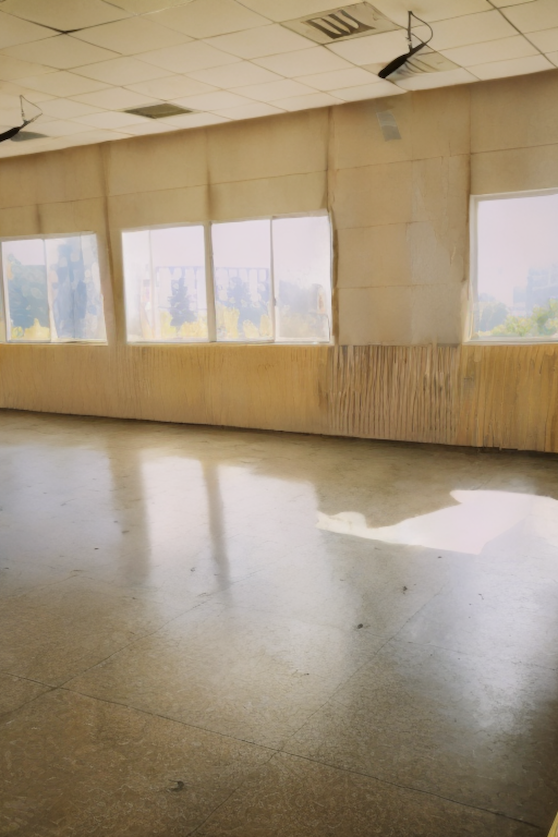} \\
        Original Image & 
        SD2~\cite{Rombach_2022_CVPR} & 
        SDXL~\cite{podell2023sdxlimprovinglatentdiffusion, noauthor_diffusersstable-diffusion-xl-10-inpainting-01_nodate}& 
        LaMa~\cite{suvorov2021resolutionrobustlargemaskinpainting}& 
        Ours
    \end{tabular}
    \caption[Qualitative Comparison of inpainting methods for in-the-wild images]{\textbf{Qualitative Comparison of inpainting methods} for in-the-wild images.}
    \label{fig:inpainting_inthewild}
\end{figure*}

SD2 Inpainting~\cite{Rombach_2022_CVPR} and SDXL Inpainting~\cite{podell2023sdxlimprovinglatentdiffusion, noauthor_diffusersstable-diffusion-xl-10-inpainting-01_nodate} rely heavily on surrounding contextual information to fill in masked areas during inpainting. Consequently, they are prone to hallucination issues, often adding unintended instances based on contextual cues like shadows. To tackle this issue, we expand the inpainting mask. Nonetheless, this adjustment could obscure key room layout indicators, such as skirting lines and wall corners. As a result, this approach can lead to inaccuracies in room layout, with altered dimensions and misaligned walls (As shown in~\cref{fig:inpainting_hypersim},~\cref{fig:inpainting_3dfutre}). Furthermore, when using SDXL Inpainting, the inpainting strength must be set below 1.0, which can result in visible artefacts appearing in the original instances within the masked area (As shown in~\cref{fig:inpainting_hypersim},~\cref{fig:inpainting_3dfutre},~\cref{fig:inpainting_inthewild}).

LaMa~\cite{suvorov2021resolutionrobustlargemaskinpainting} propagates the neighbouring textures without clearly defined boundaries, leading to blurred inpainted areas that can obscure critical details like skirting lines and wall edges. This blurring effect can also introduce unwanted elements, such as dark shadows, which further degrade the visual quality of the inpainted regions.

Thanks to the random dilation or erosion of inpainting masks during training, our method can naturally handle imprecise inpainting masks without the need to enlarge them. This strategy facilitates straightforward room layout prediction without mask engineering and allows for more precise removal of foreground instances in case of imperfect masks. By preserving structural integrity, our method maintains a high degree of coherence with the original scene, with reduced artefacts and distortions in the room’s geometry. Additionally, during training, unlike SD2 Inpainting~\cite{Rombach_2022_CVPR} and SDXL Inpainting~\cite{podell2023sdxlimprovinglatentdiffusion, noauthor_diffusersstable-diffusion-xl-10-inpainting-01_nodate}, where the outside-of-mask areas remain unchanged, our method allows for adjusting these regions based on contextual information. This enables our method to remove shadows and restore the scene with normal lighting conditions, resulting in a more natural appearance.

\section{Amodal Completion}

\begin{figure*}
    \centering    
    \setlength{\wid}{0.25\textwidth}
    \setlength{\mrg}{0cm}
    \begin{tabular}{@{}cccc@{}}
        \includegraphics[width=\wid]{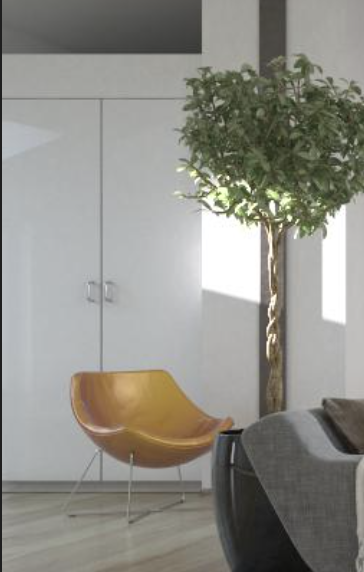} &
        \hspace{-3mm}\includegraphics[width=\wid]{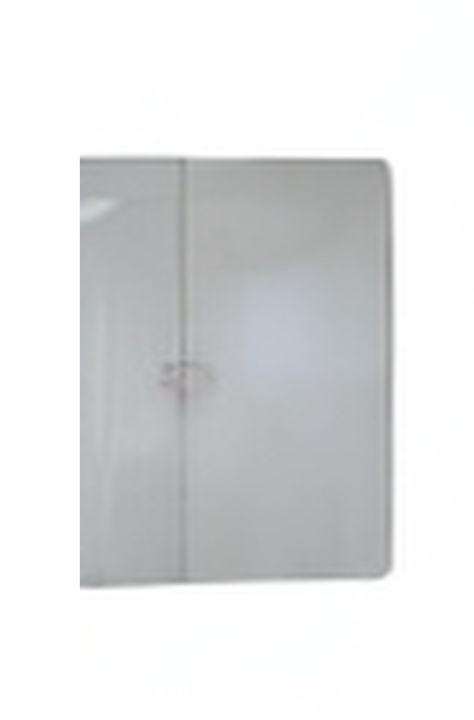} &
        \hspace{-3mm}\includegraphics[width=\wid]{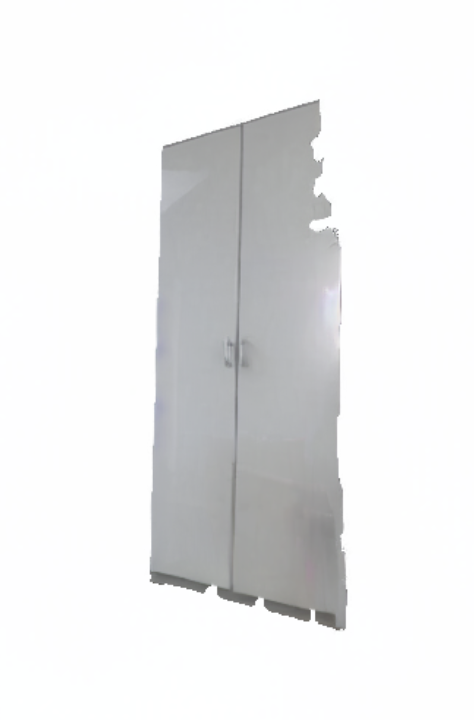} &
        \hspace{-3mm}\includegraphics[width=\wid]{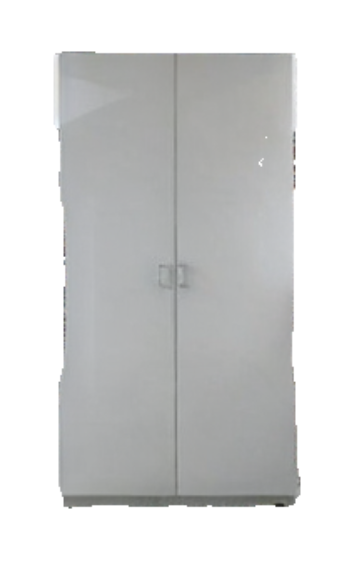} 
        \\ 
        \includegraphics[width=\wid]{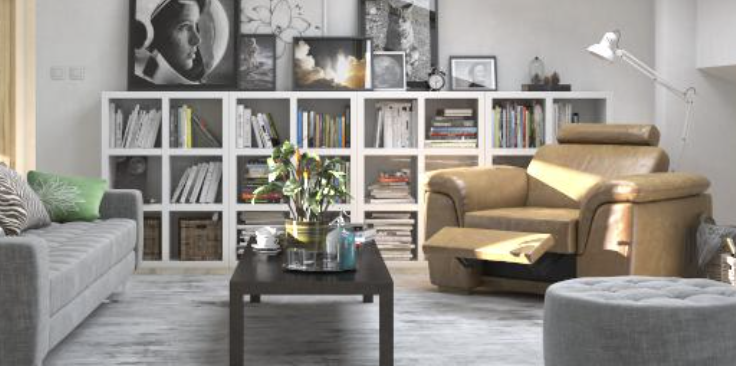} &
        \hspace{-3mm}\includegraphics[width=\wid]{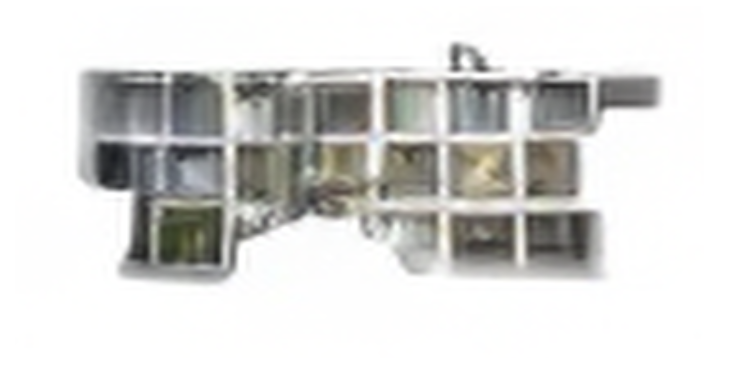} &
        \hspace{-3mm}\includegraphics[width=\wid]{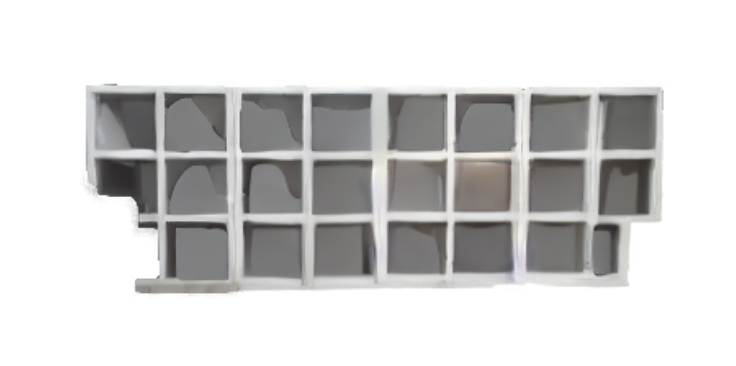} &
        \hspace{-3mm}\includegraphics[width=\wid]{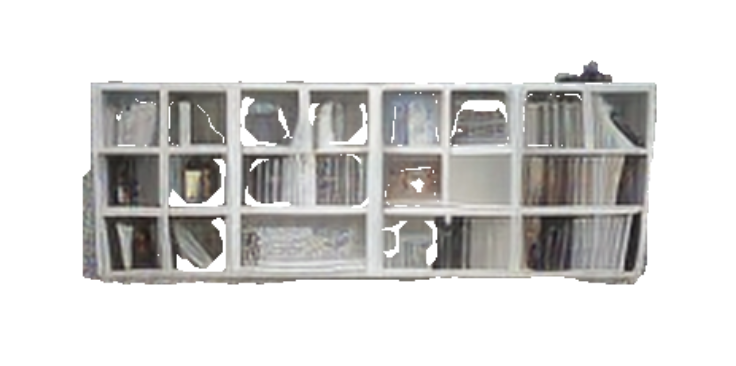} 
        \\ 
        \includegraphics[width=\wid]{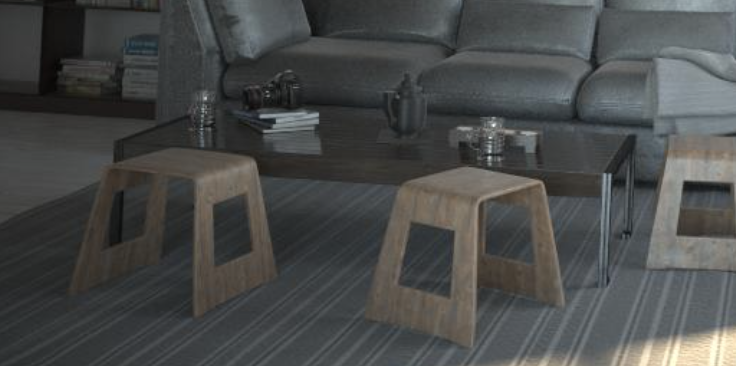} &
        \hspace{-3mm}\includegraphics[width=\wid]{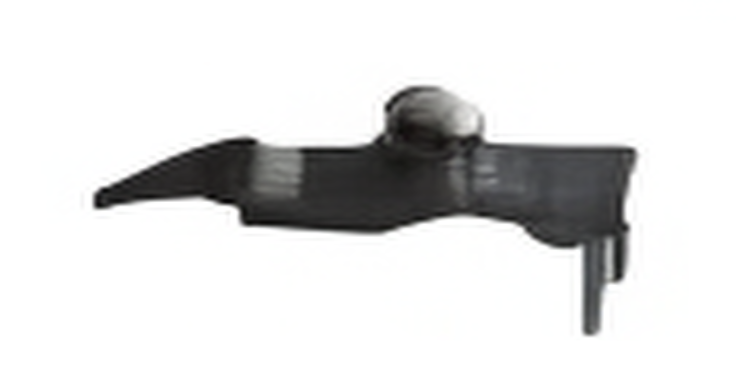} &
        \hspace{-3mm}\includegraphics[width=\wid]{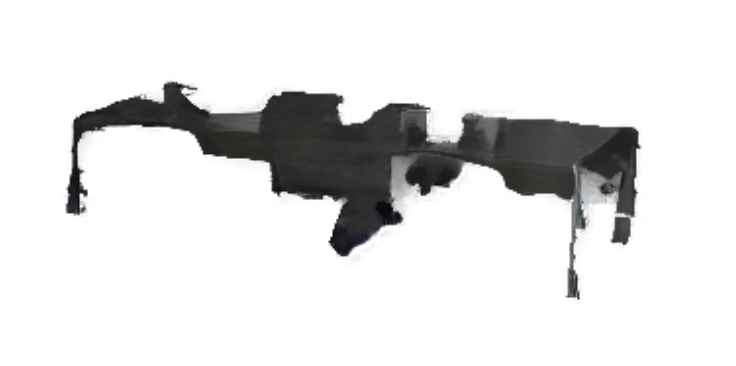} &
        \hspace{-3mm}\includegraphics[width=\wid]{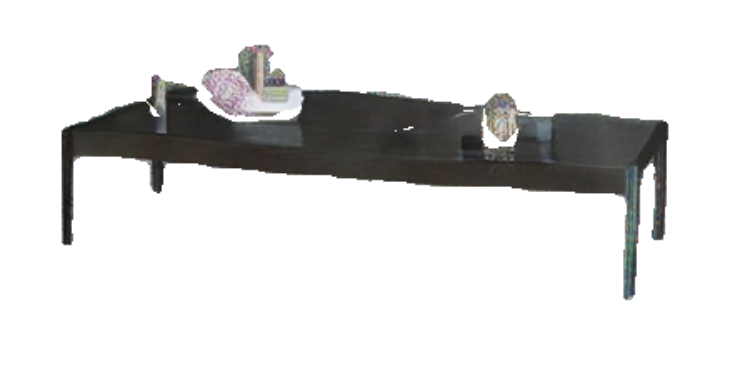} 
        \\ 
        \includegraphics[width=\wid]{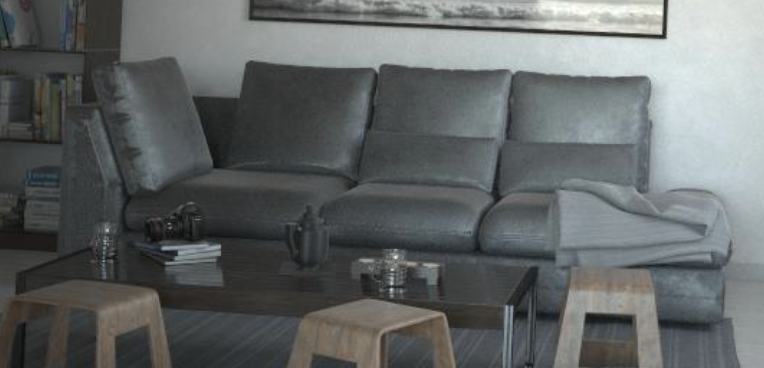} &
        \hspace{-3mm}\includegraphics[width=\wid]{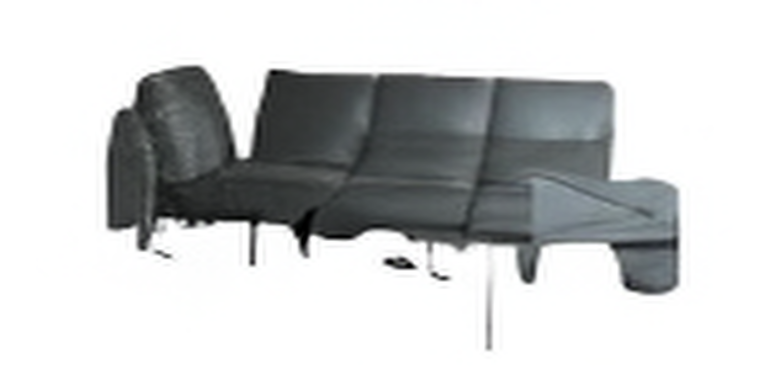} &
        \hspace{-3mm}\includegraphics[width=\wid]{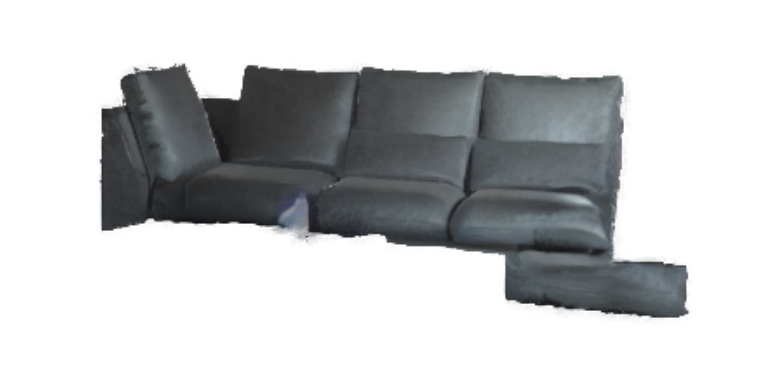} &
        \hspace{-3mm}\includegraphics[width=\wid]{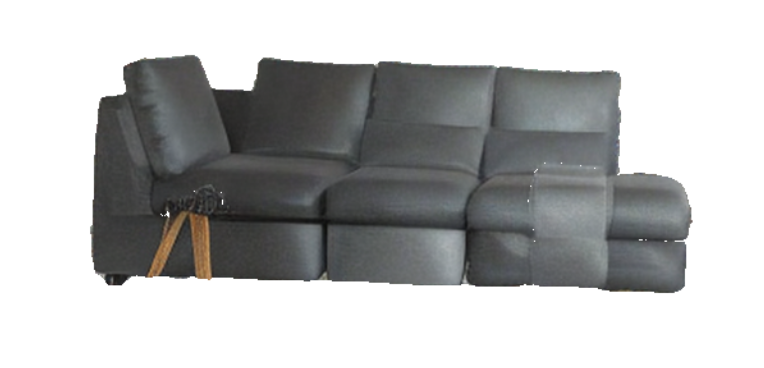} 
        \\ 
        Original Image & 
        Pix2gestalt~\cite{ozguroglu_pix2gestalt_2024} & 
        Gen3DSR~\cite{dogaru_generalizable_2024}& 
        Ours
    \end{tabular}
    \caption[Qualitative Comparison of our Amodal Completion module]{\textbf{Qualitative Comparison of our Amodal Completion module.} Compared with Pix2gestalt~\cite{ozguroglu_pix2gestalt_2024} and Gen3DSR~\cite{dogaru_generalizable_2024}, even though both methods are trained on specific datasets, our method consistently achieves better results in handling occlusion and out-of-frame cases.}
    \label{fig:amodal}
\end{figure*}
\subsection{Evaluation}
For amodal completion tasks, methods typically rely on training with hand-crafted datasets containing ground truth annotations to restore occluded or out-of-frame parts of objects. However, creating such datasets is costly and limits the generalization capability of these methods to unseen objects. Our lightweight amodal completion approach bypasses the need for additional training on specific datasets. Instead, we utilize SD2 inpainting~\cite{Rombach_2022_CVPR}, combined with carefully engineered masks, to restore occluded or out-of-frame parts of indoor instances. In the absence of an indoor-adapted dataset with ground truth for amodal completion, we conduct a qualitative evaluation of our model's performance, comparing it against Pix2gestalt~\cite{ozguroglu_pix2gestalt_2024} and the amodal completion module in Gen3DSR~\cite{dogaru_generalizable_2024}, both of which are trained on a specific dataset.

As illustrated in~\cref{fig:amodal}, Pix2gestalt performs the worst in amodal completion for indoor instances as it lacks specialized training. While Gen3DSR shows improved results due to its training on occluded indoor objects, it still faces challenges in guessing the shape boundaries of the processed instances. Our method, though in a zero-shot setting, consistently delivers the best performance owing to well-engineered inpainting masks.

\section{Single-View Room Reconstruction}
\subsection{Evaluation}
We evaluate our method's performance on single-view 3D reconstruction across various indoor scene types. For quantitative evaluation, we choose the 3D-Front dataset~\cite{fu20213d}, a synthetic indoor scenes dataset with a large number of rooms populated by high-quality textured 3D models. We manually select 100 scenes to ensure they cover a diverse range of room categories and correspond to meaningful, natural indoor scenes. These selections were made to also avoid instance superposition, unrealistic viewpoint settings and repetitive scene layouts. For qualitative evaluation, we further include the SUN RGB-D dataset to test the effectiveness and generalization capabilities of our method in real-world indoor scene settings.

\subsubsection{Room Background Reconstruction}
\begin{table}[b]
\centering
\begin{tabular}{l@{\hspace{50pt}}c}
\toprule
\textbf{Method} & \textbf{CD $\downarrow$} \\ 
\midrule
PlaneRCNN~\cite{liu_planercnn_2019} & 0.717 \\
InstPIFu~\cite{liu2022towards} & 0.481 \\
Gen3DSR~\cite{dogaru_generalizable_2024} & 0.303 \\
\textbf{Ours} & \textbf{0.103} \\
\bottomrule
\end{tabular}
\caption[Quantitative Evaluation of Room Background Reconstruction on the 3D-FRONT Dataset~\cite{fu20213d}]{\textbf{Quantitative Evaluation of Room Background Reconstruction} on the 3D-FRONT Dataset~\cite{fu20213d}.}
\label{tab:background_comparison}
\end{table}
We start by evaluating the quality of the reconstructed room background. Layout IoU (Intersection over Union) is typically used for assessing the quality of layout reconstruction, as it measures the overlap between the predicted and ground truth layouts. However, our reconstruction is obtained from a monocular image, which results in partial and incomplete views of the room, the IoU may not adequately capture the differences between the predicted and actual layouts as it primarily focuses on the overall shape overlap and is less effective in partial reconstructions. To address its limitation, we decide to use Chamfer distance~\cite{barrow1977parametric} to measure the average closest-point distance between 10K points sampled from the predicted and ground truth 3D models aligned using ICP~\cite{besl1992method}.

We compare our method with three methods: the specialized plane reconstruction network PlaneRCNN~\cite{liu_planercnn_2019}, as well as the implicit surface-based methods InstPIFu~\cite{liu2022towards} and Gen3DSR~\cite{dogaru_generalizable_2024}. As shown in Table~\ref{tab:background_comparison}, earlier CNN-based methods like PlaneRCNN have difficulty handling complex indoor scenes, leading to the worst performance. InstPIFu and Gen3DSR employ implicit functions to model room backgrounds based on visible points, its limited extrapolation capability with signed distance functions frequently results in uneven surfaces, causing inaccuracies such as unrealistic spatial configurations in the reconstructed room model. While Gen3DSR achieves better results due to its more accurate depth estimation, it often has difficulty identifying room background elements during 2D semantic understanding, leading to issues such as background holes as shown in~\cref{fig:3dfront_qua2}. Instead of attempting to directly predict the room background in 3D from incomplete information, our method first estimates the unseen areas in 2D before reconstructing the full geometry in 3D, which greatly simplifies the problem. This two-step approach substantially enhances the quality of the reconstruction, making it the only method that consistently achieves both high visual quality and geometric fidelity.

\subsubsection{Holistic Scene Reconstruction}
For the holistic scene reconstruction, we compare our method quantitatively against SOTA methods in single-view indoor scene 3D reconstruction, including Gen3DSR~\cite{dogaru_generalizable_2024} and InstPIFu~\cite{liu2022towards}. To access the fidelity of our 3D scene reconstruction, in addition to using Chamfer Distance~\cite{barrow1977parametric} to evaluate the geometric accuracy of the reconstructed model, we employ F-score~\cite{knapitsch_tanks_2017} to evaluate the completeness and correctness of the reconstruction. The F-score provides a balanced measure of precision, the proportion of reconstructed points that are close to the ground truth, and recall, the extent to which the ground truth points are captured by the reconstruction. Both metrics are computed on 10K points sampled from the reconstructed scene mesh and the ground truth scene mesh aligned using ICP~\cite{besl1992method}. For the visual comparison, we compare our method against InstPIFu~\cite{liu2022towards}, Gen3DSR~\cite{dogaru_generalizable_2024}. We also include Total3D~\cite{nie_total3dunderstanding_2020} in our comparisons for its reconstruction of foreground instances. Please note that no quantitative scores for holistic scene reconstruction are reported for Total3D~\cite{nie_total3dunderstanding_2020} due to its over-simplified modelling of the room background as a cuboid.

\begin{table}[b]
\centering
\begin{tabular}{lcc}
\toprule
\textbf{Method} & \textbf{CD $\downarrow$} & \textbf{F-Score $\uparrow$} \\
\midrule
InstPIFu~\cite{liu2022towards} & 0.119 & 70.63 \\
Gen3DSR~\cite{dogaru_generalizable_2024} & 0.182 & 65.35 \\
Ours & \textbf{0.097} & \textbf{80.96} \\
\bottomrule
\end{tabular}
\caption[Quantitative Evaluation of Holistic Scene Reconstruction on the 3D-FRONT dataset~\cite{fu20213d}.]{\textbf{Quantitative Evaluation of Holistic Scene Reconstruction} including background and foreground elements on the 3D-FRONT dataset~\cite{fu20213d}. Our method outperforms both baselines in terms of both Chamfer Distance and F-Score.}
\label{tab:quantitative3dfont}
\end{table}

\begin{figure*}[htb]
    \centering
    \setlength{\wid}{0.195\textwidth}
    \setlength{\mrg}{-0.2cm}
    \begin{tabular}{@{}ccccc@{}}
        \includegraphics[width=\wid]{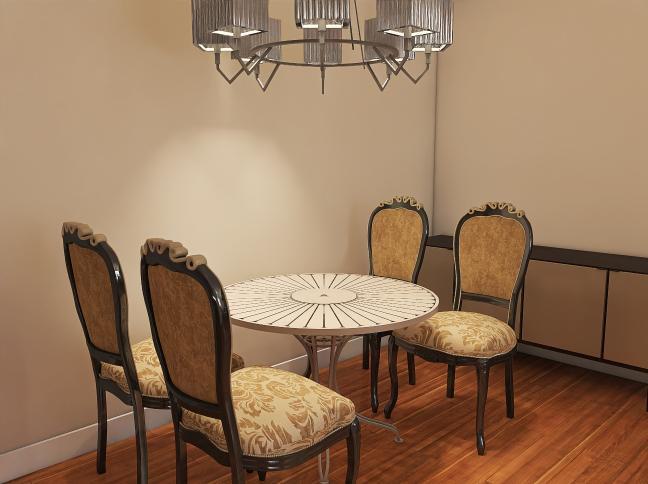} &
        \hspace{-3mm}\includegraphics[width=\wid]{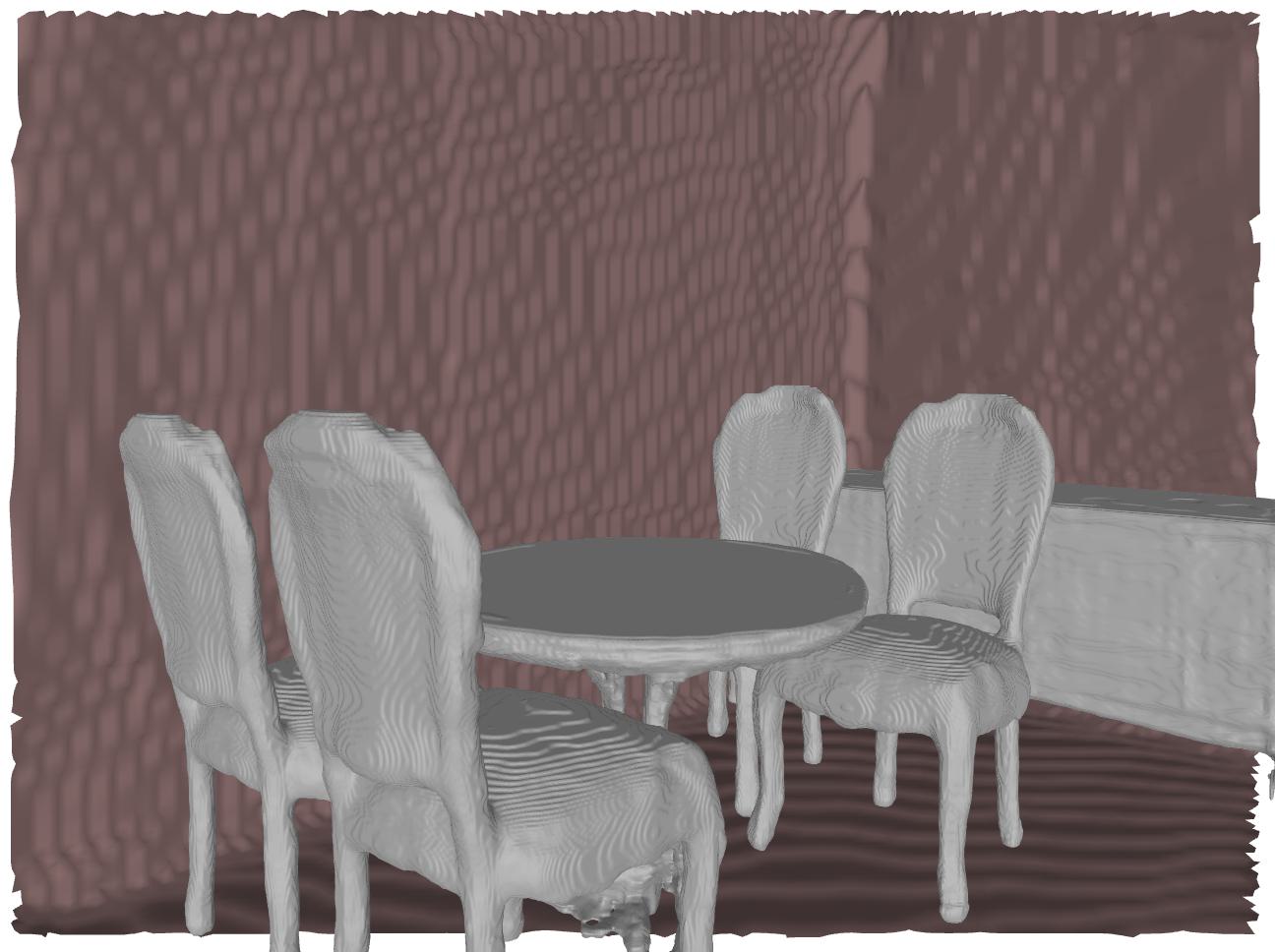} &
        \hspace{-3mm}\includegraphics[width=\wid]{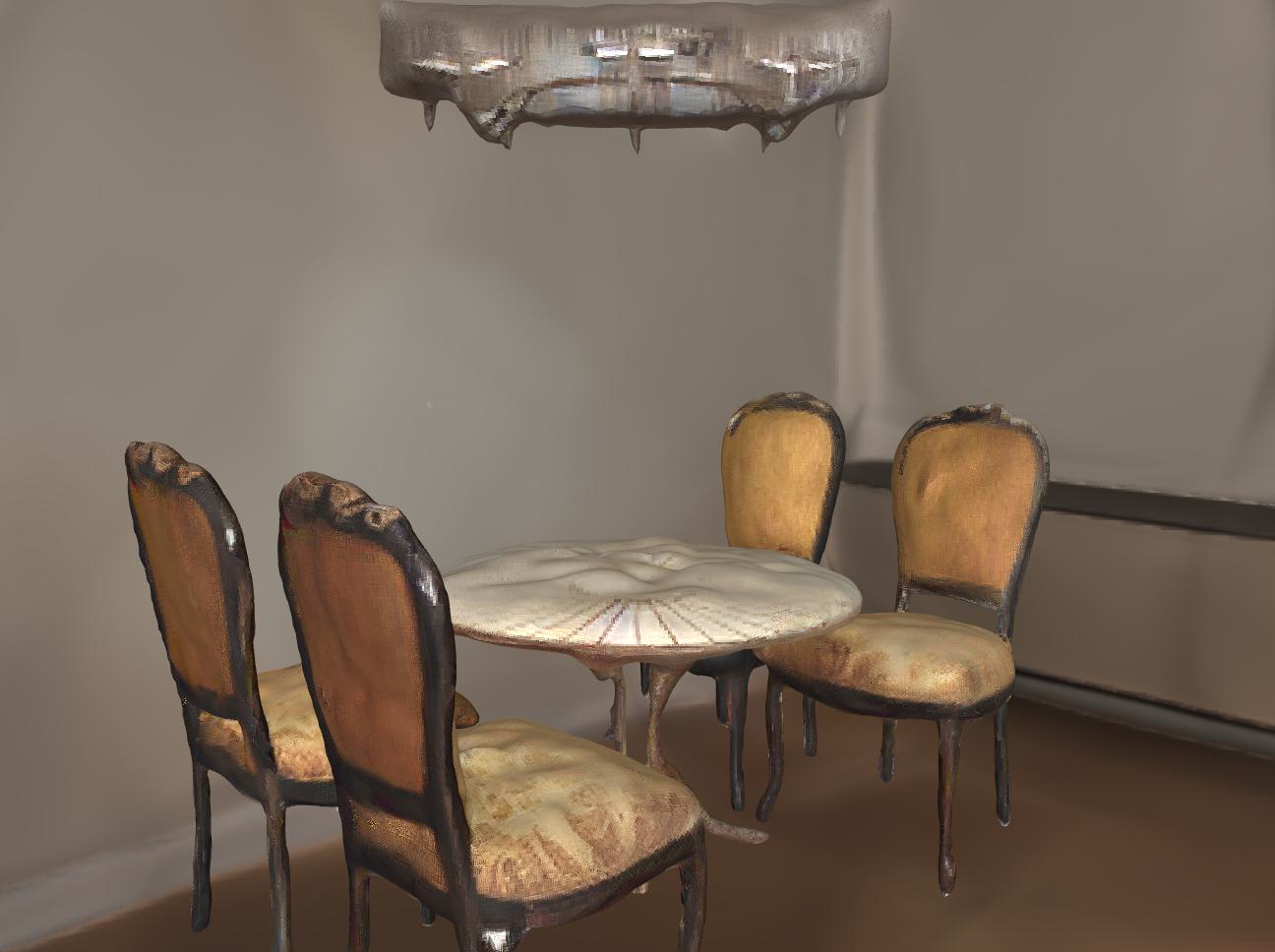} &
        \hspace{-3mm}\includegraphics[width=\wid]{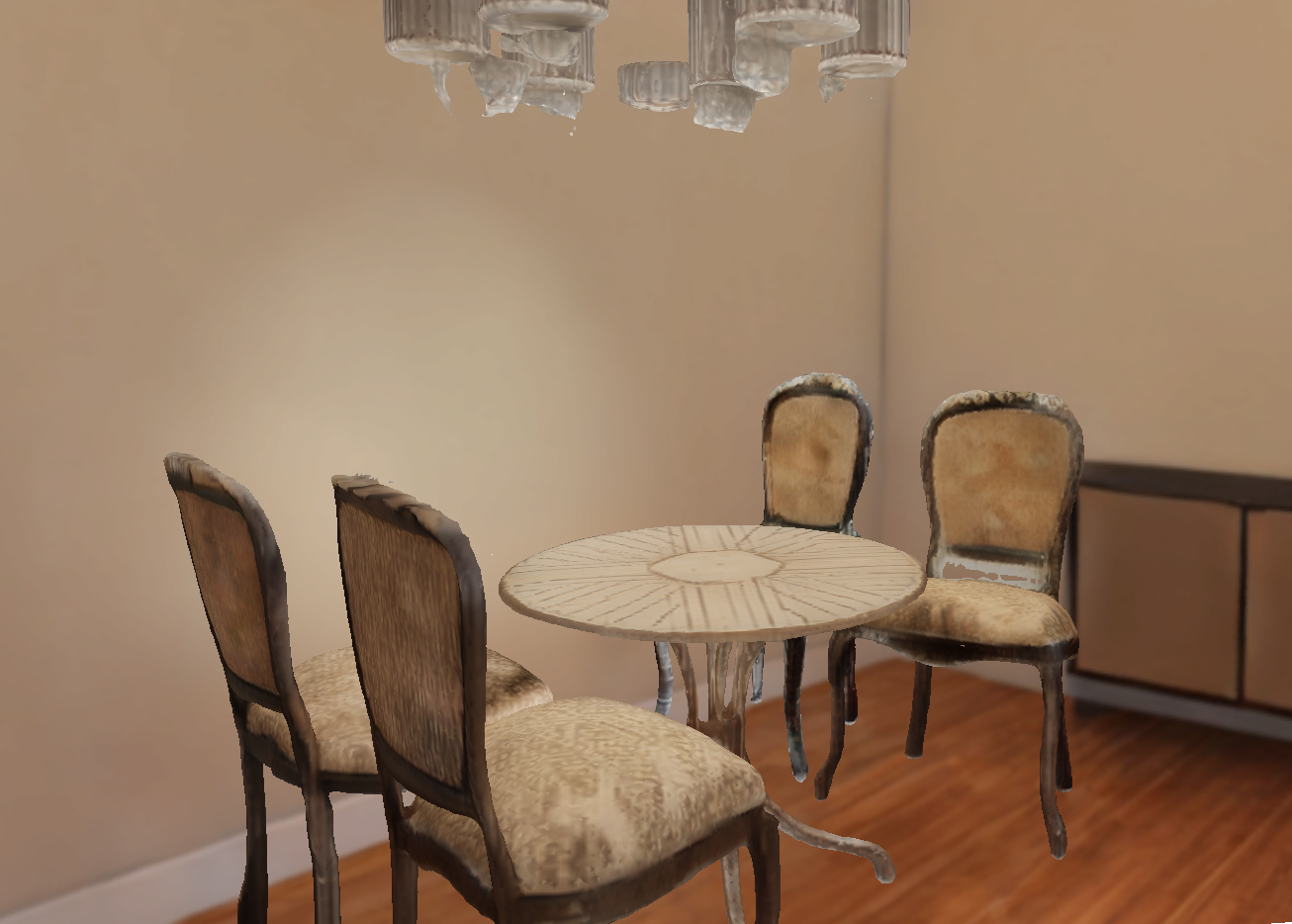} & 
        \hspace{-3mm}\includegraphics[width=\wid]{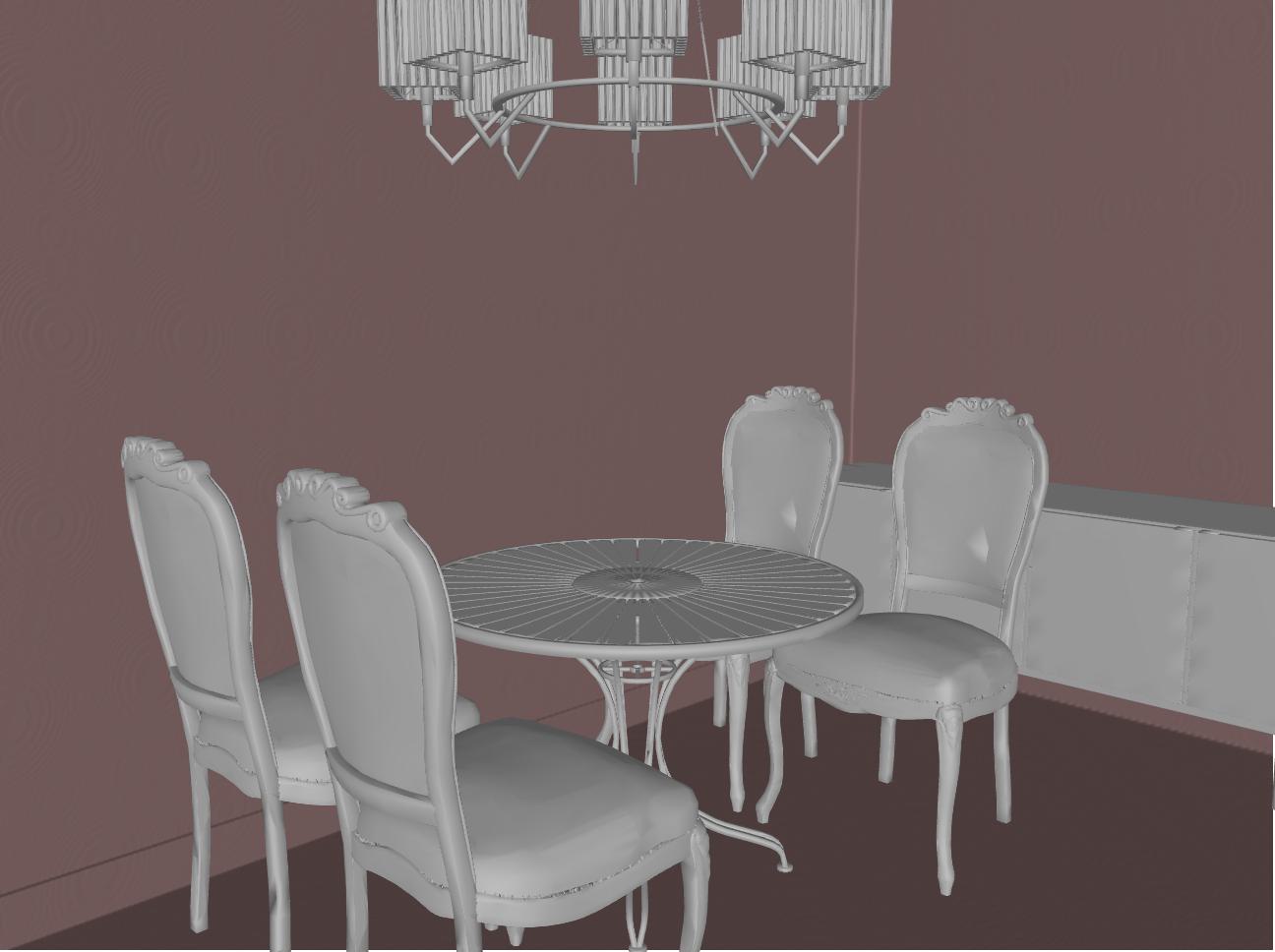}
        \\ 
         &
        \includegraphics[width=\wid]{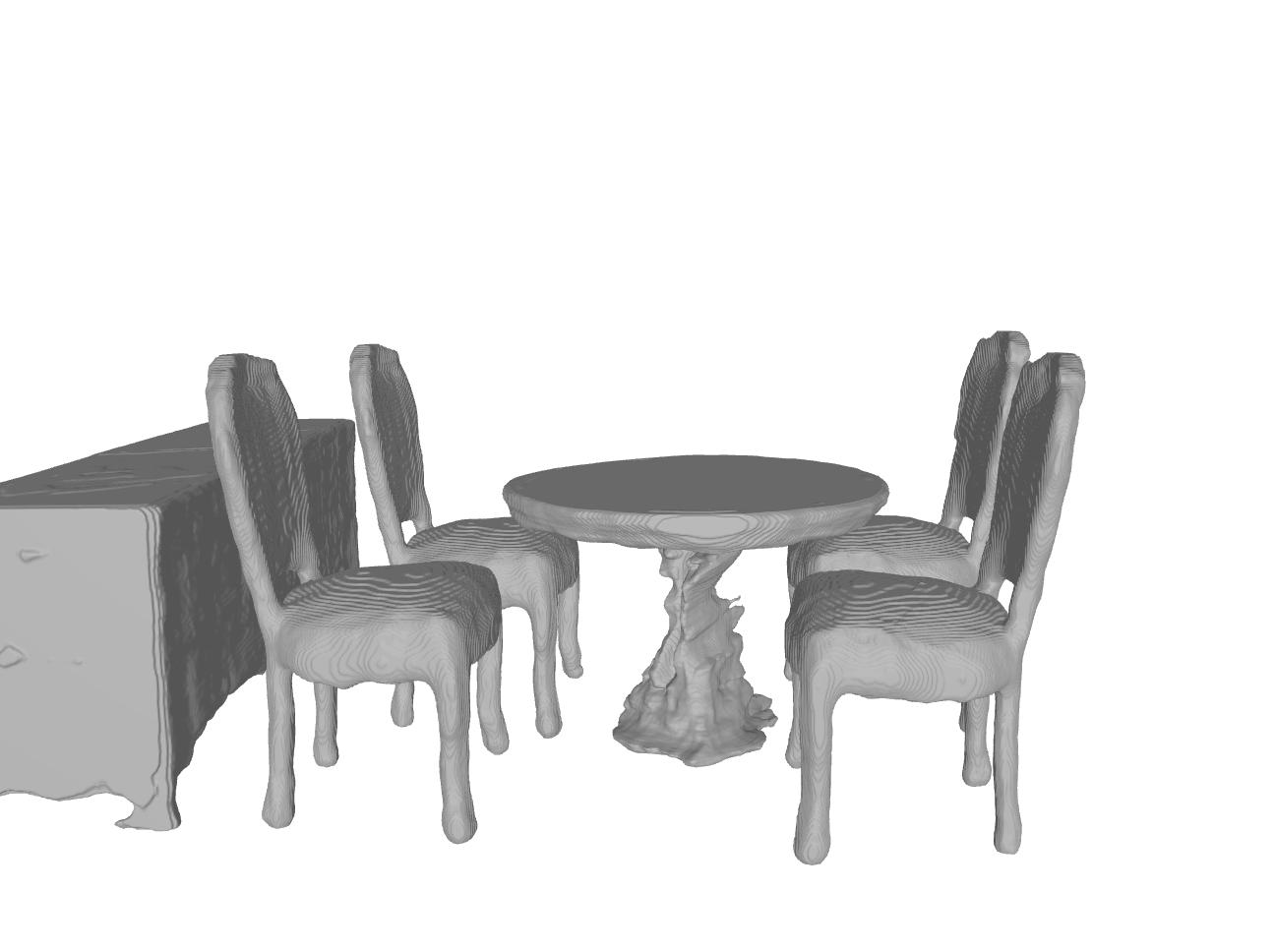} &
        \hspace{-3mm}\includegraphics[width=\wid]{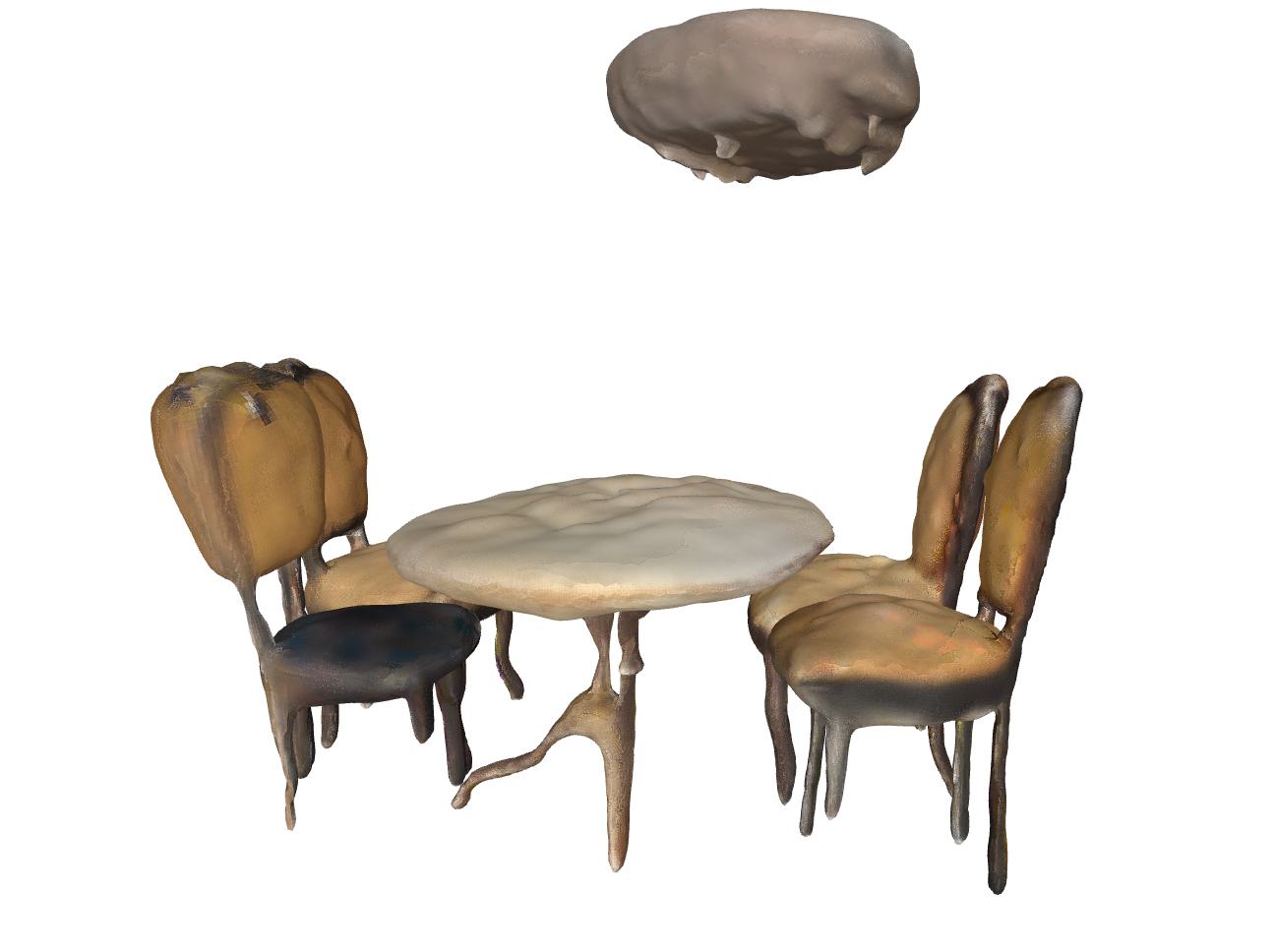} & 
        \hspace{-3mm}\includegraphics[width=\wid]{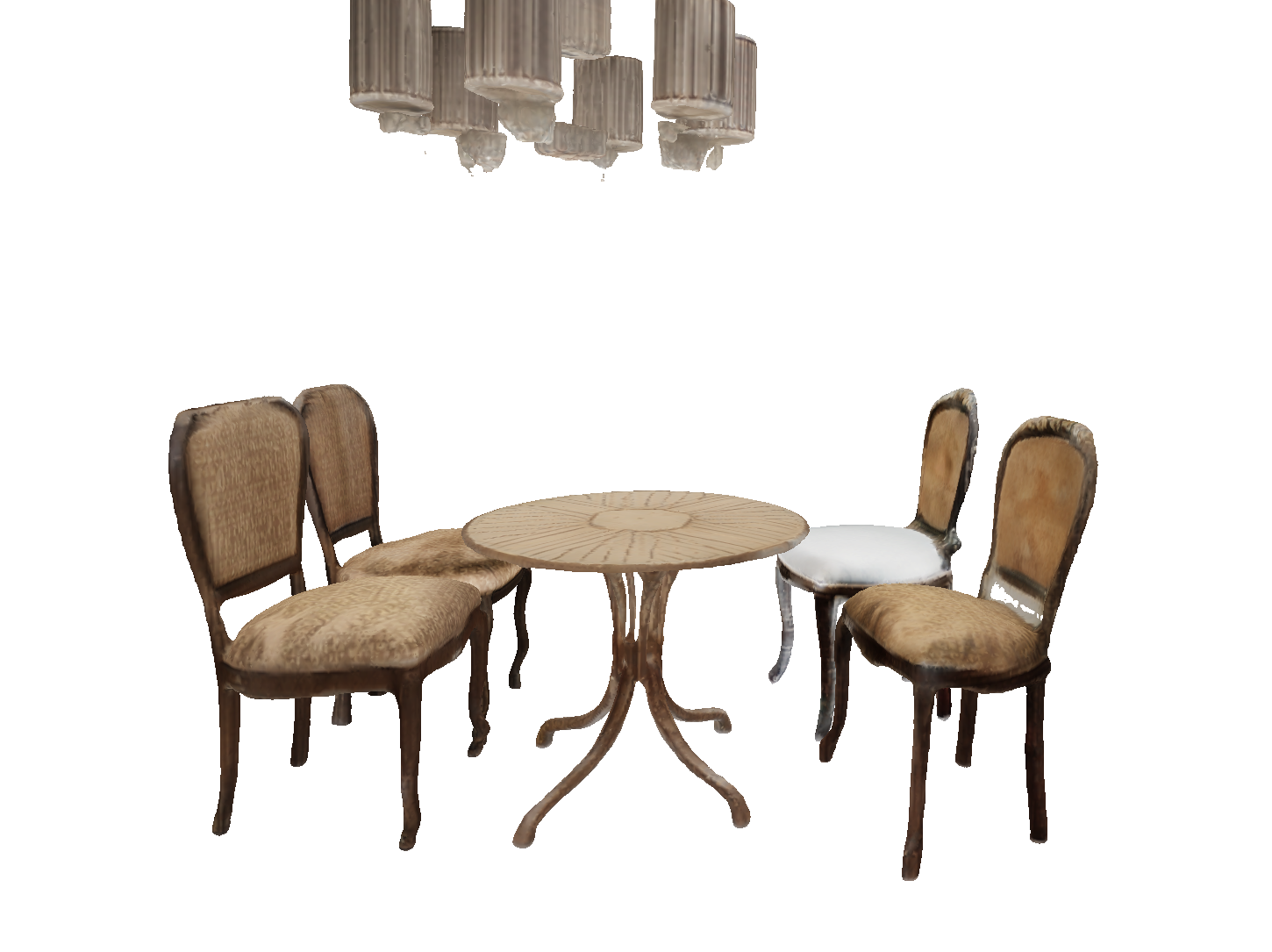} & 
        \hspace{-3mm}\includegraphics[width=\wid]{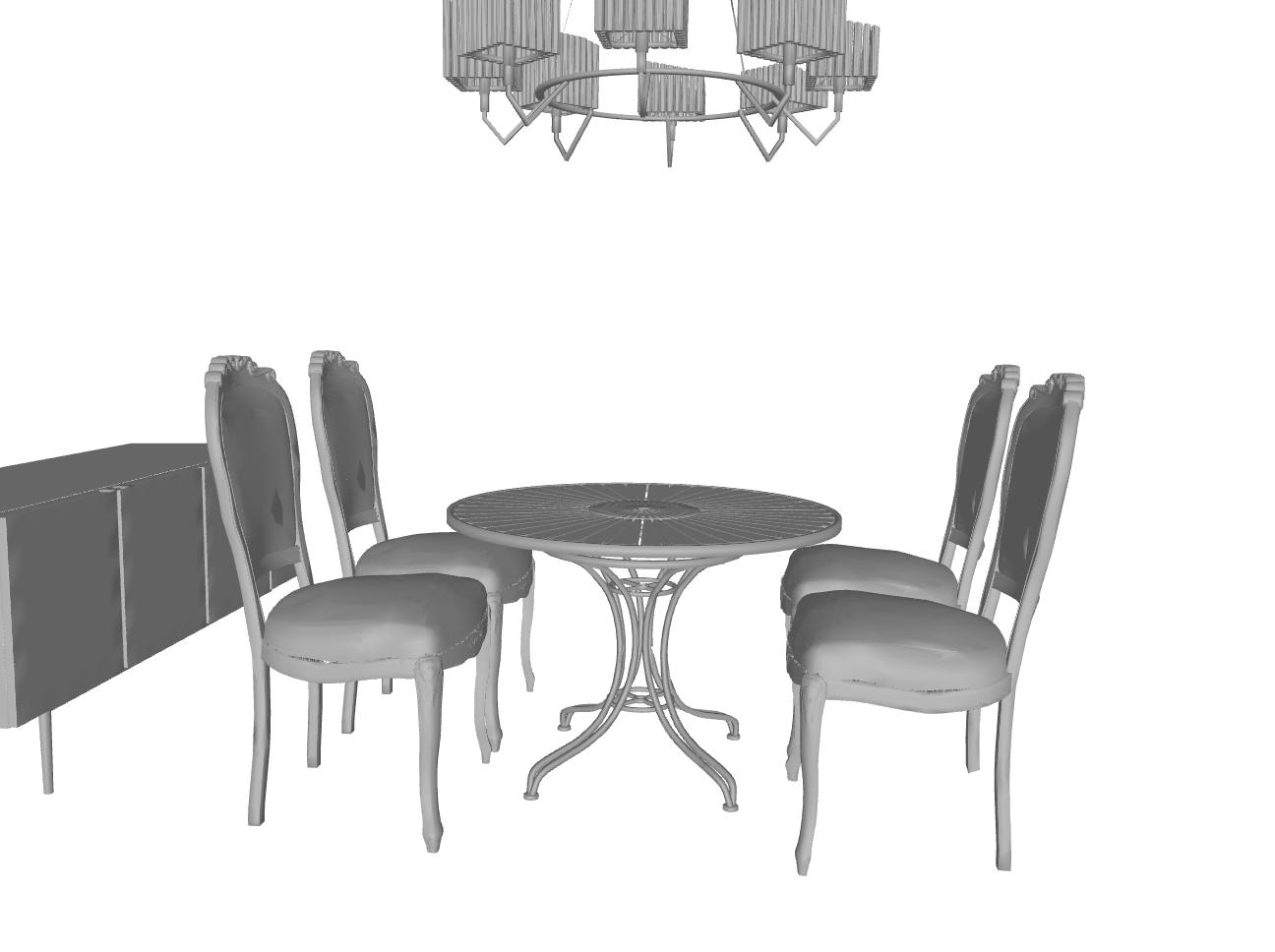} \\
        \includegraphics[width=\wid]{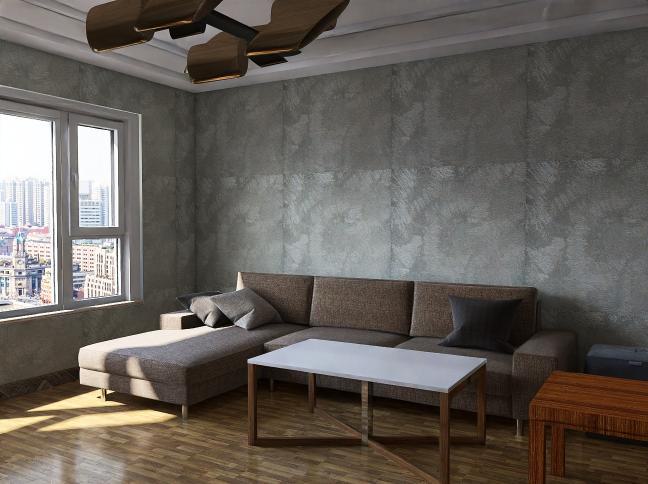} &
        \hspace{-3mm}\includegraphics[width=\wid]{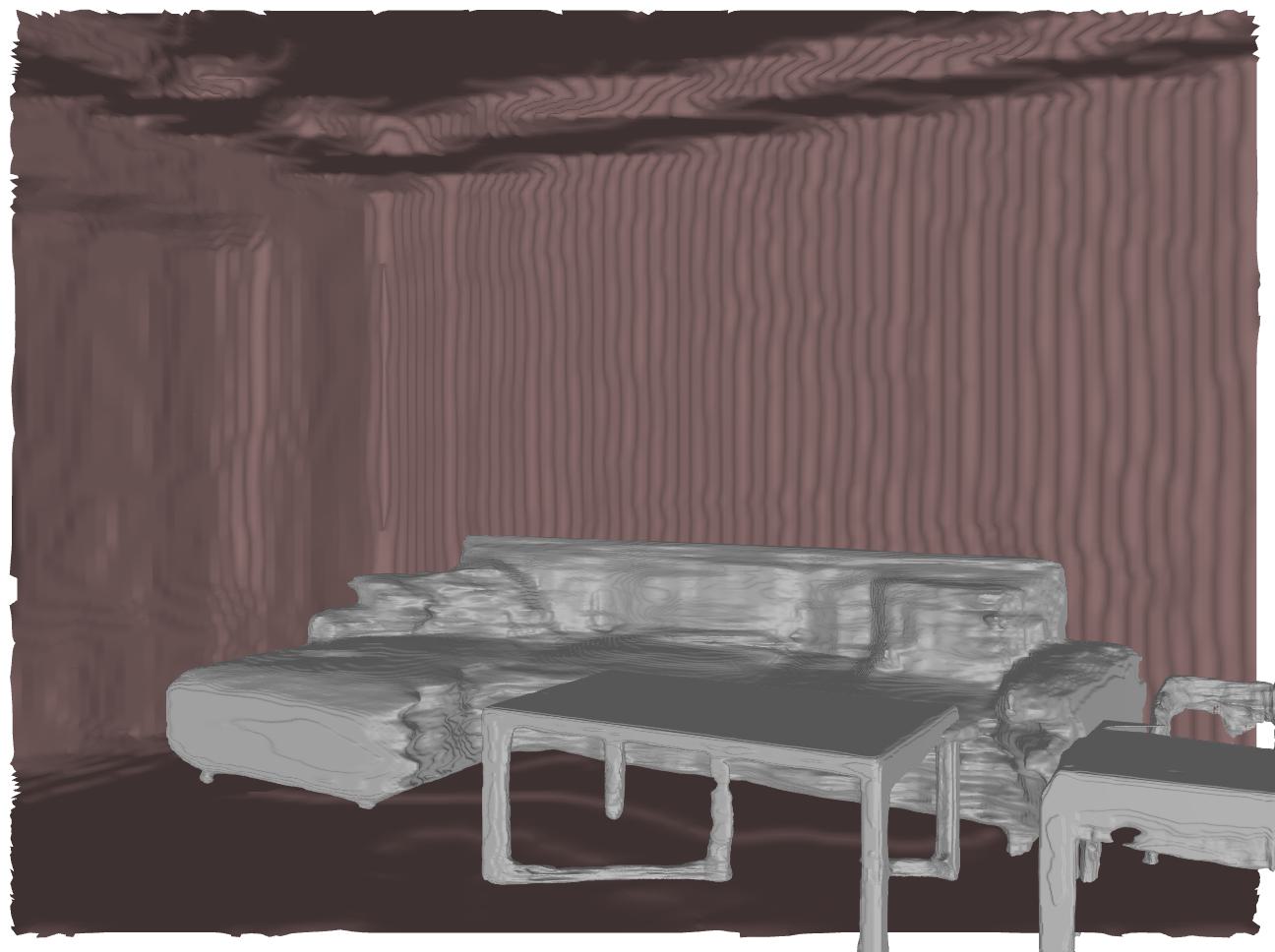} &
        \hspace{-3mm}\includegraphics[width=\wid]{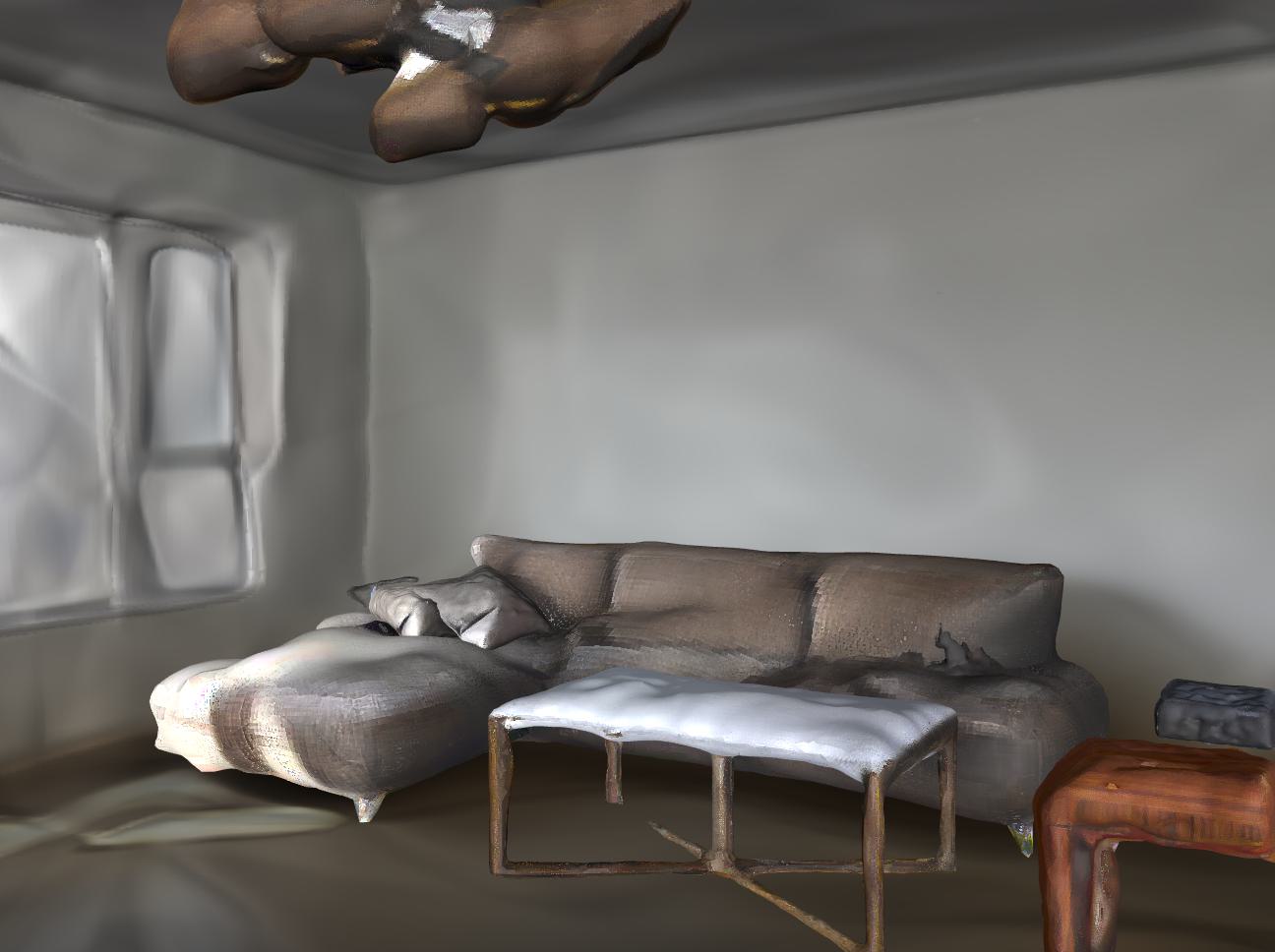} & 
        \hspace{-3mm}\includegraphics[width=\wid]{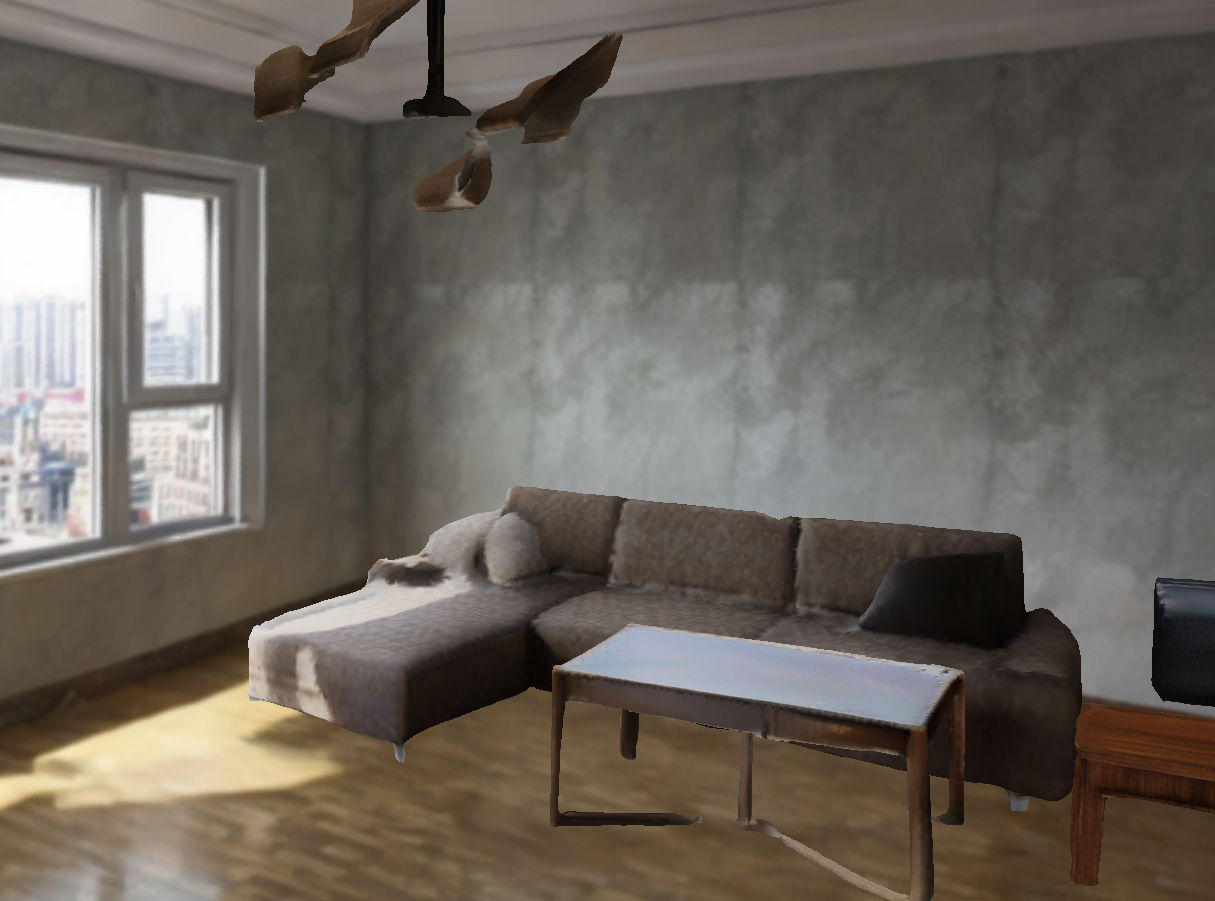} & 
        \hspace{-3mm}\includegraphics[width=\wid]{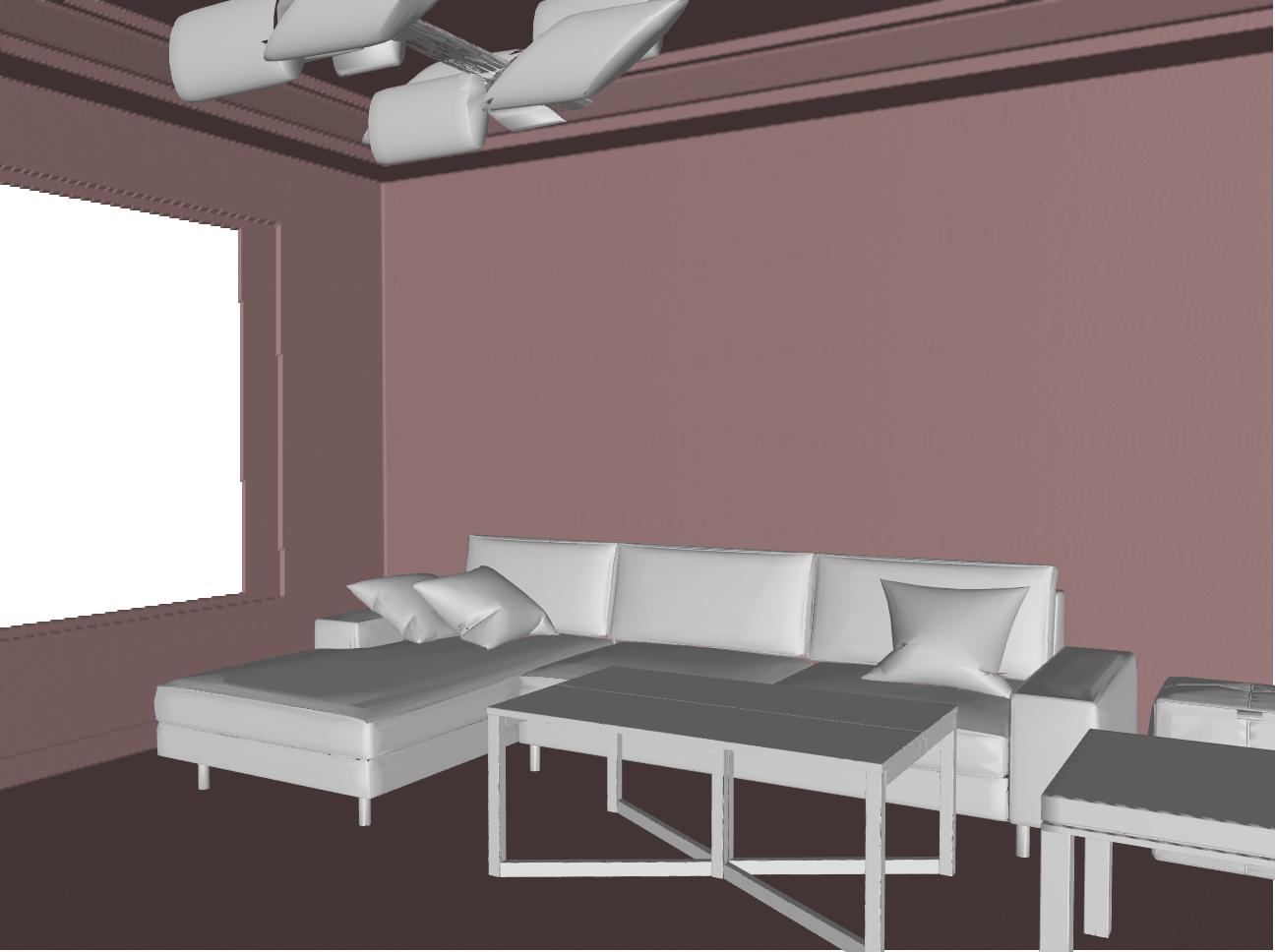} \\
         &
        \includegraphics[width=\wid]{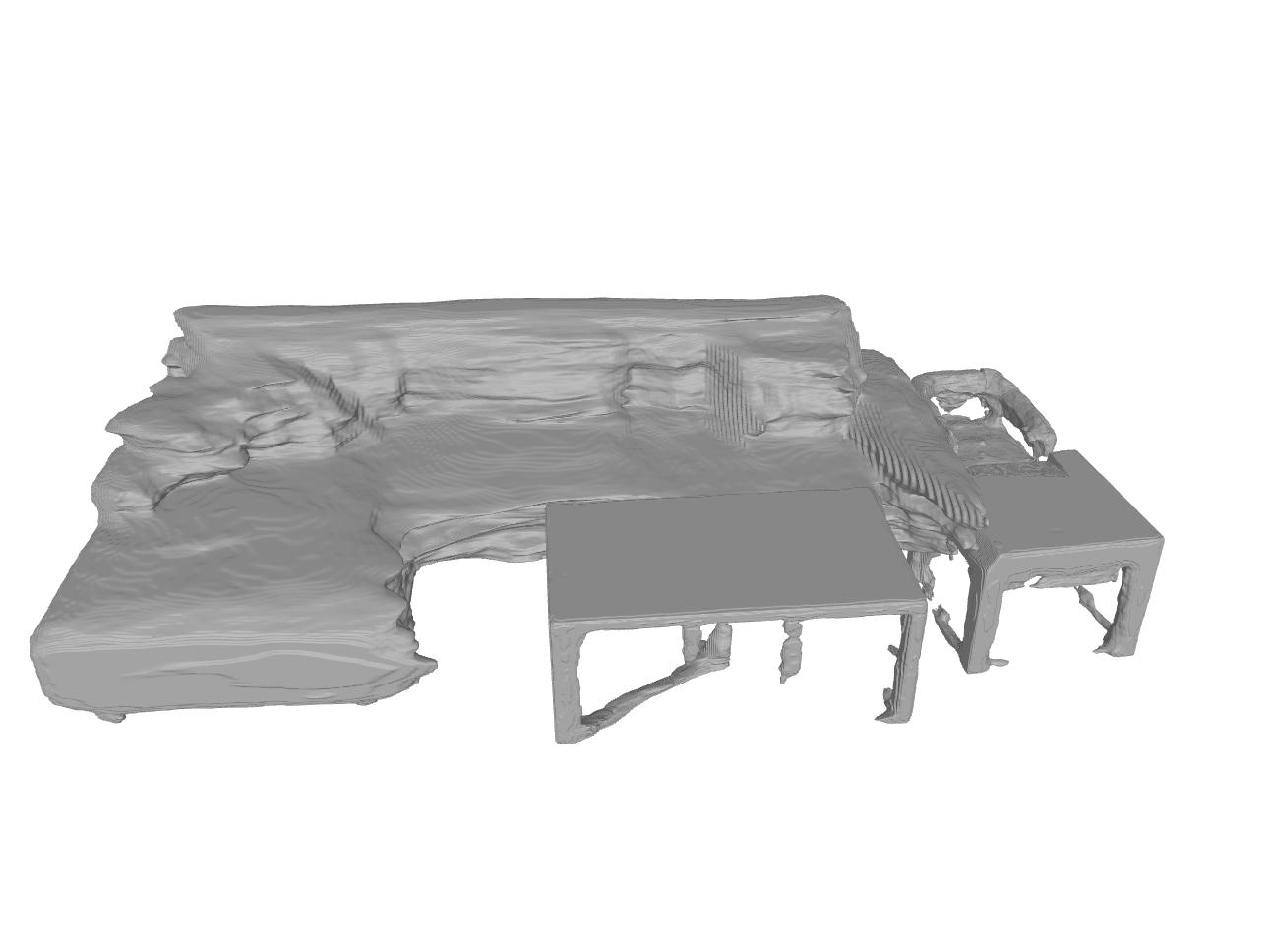} &
        \hspace{-3mm}\includegraphics[width=\wid]{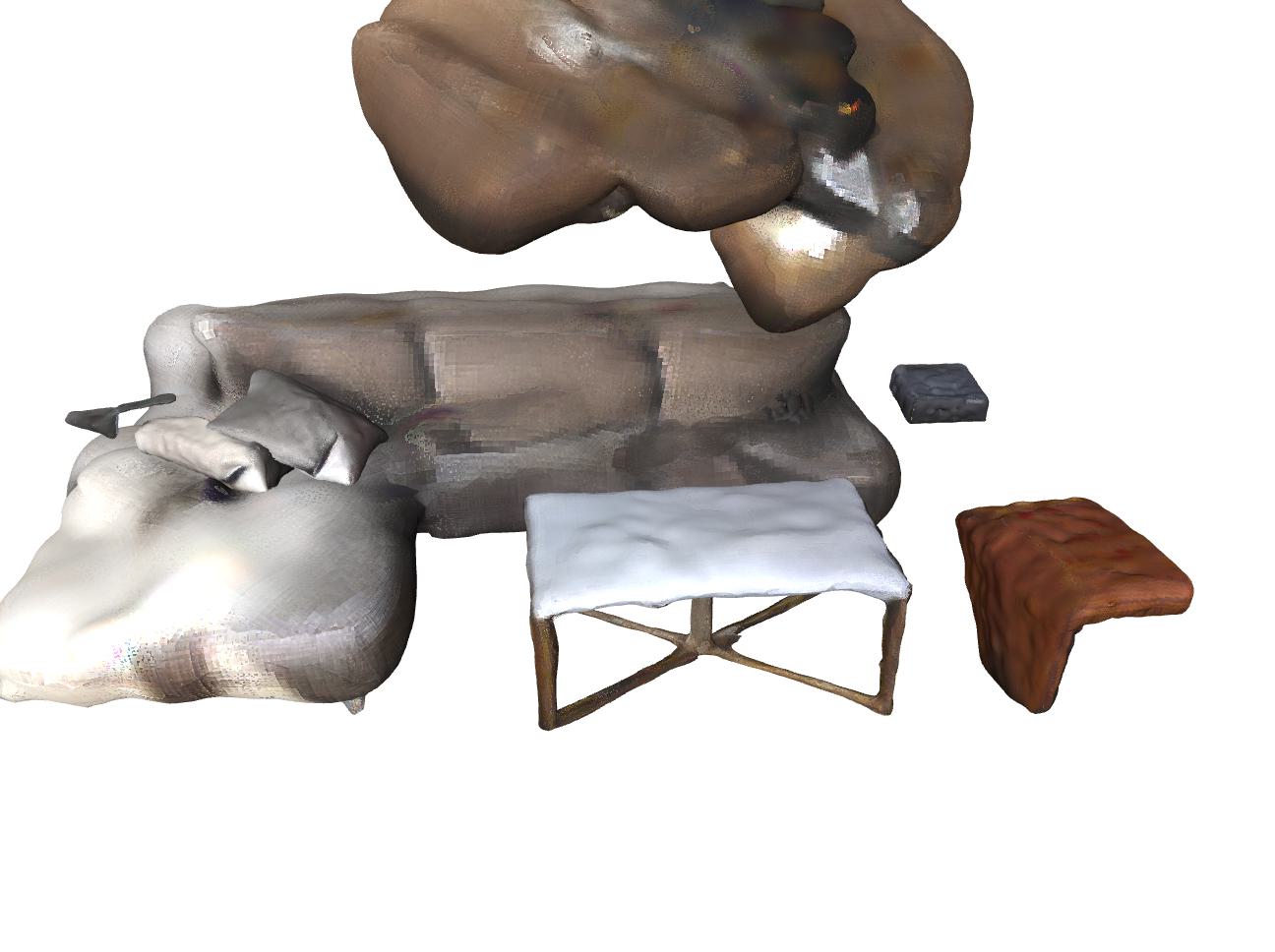} & 
        \hspace{-3mm}\includegraphics[width=\wid]{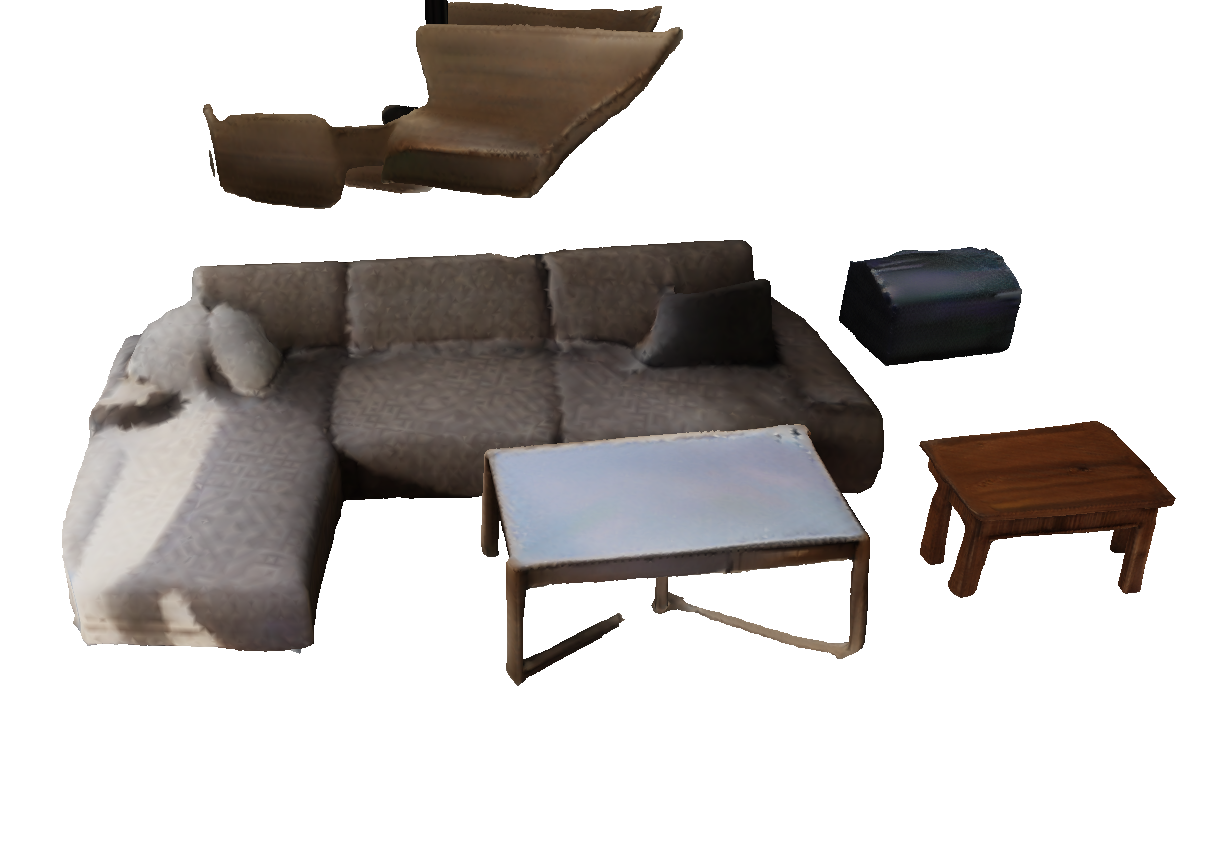} & 
        \hspace{-3mm}\includegraphics[width=\wid]{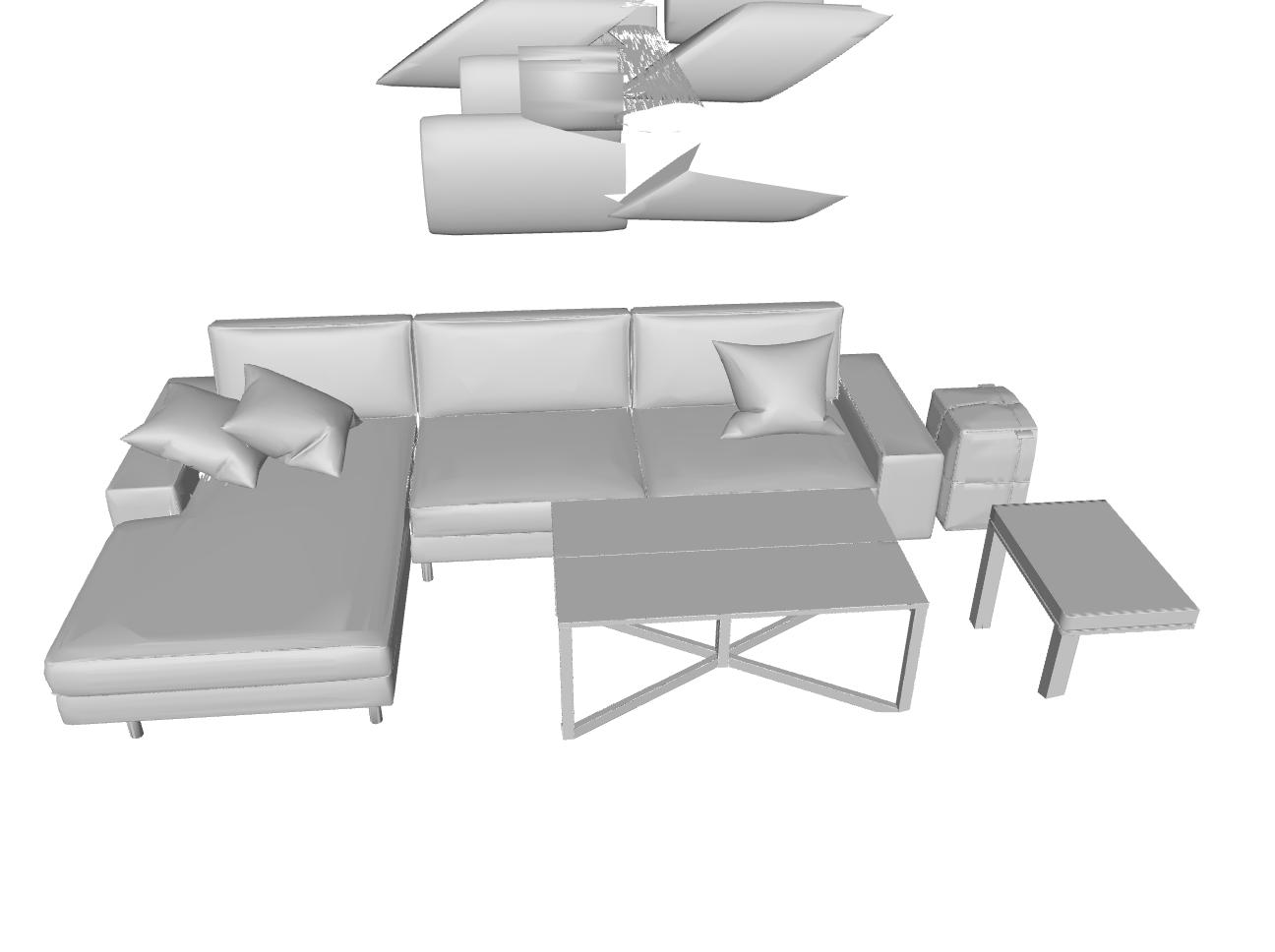} \\
        Input image & 
        InstPIFu~\cite{liu2022towards} & 
        Gen3DSR~\cite{dogaru_generalizable_2024} & 
        Ours & 
        GT geometry
    \end{tabular}
    \caption[Qualitative Comparison of Holistic Scene Reconstruction on the 3D-FRONT dataset~\cite{fu20213d} with novel view]{\textbf{Qualitative Comparison of Holistic Scene Reconstruction} on the 3D-FRONT dataset~\cite{fu20213d} for holistic scene reconstrcution with novel view. Our reconstruction remains faithful to GT geometry even from a different viewpoint than the input view.} 
    \label{fig:3dfront_qua1}
    \vspace{-0.3cm}
\end{figure*}

\begin{figure*}[t]
    \centering    
    \setlength{\wid}{0.195\textwidth}
    \setlength{\mrg}{-0.2cm}
    \begin{tabular}{@{}ccccc@{}}
        \includegraphics[width=\wid]{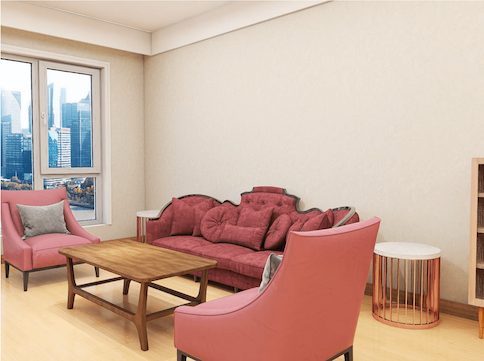} &
        \hspace{-3mm}\includegraphics[width=\wid]{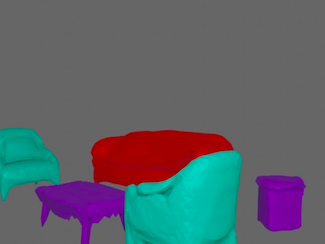}&
        \hspace{-3mm}\includegraphics[width=\wid]{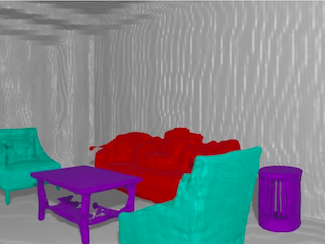} & 
        \hspace{-3mm}\includegraphics[width=\wid]
        {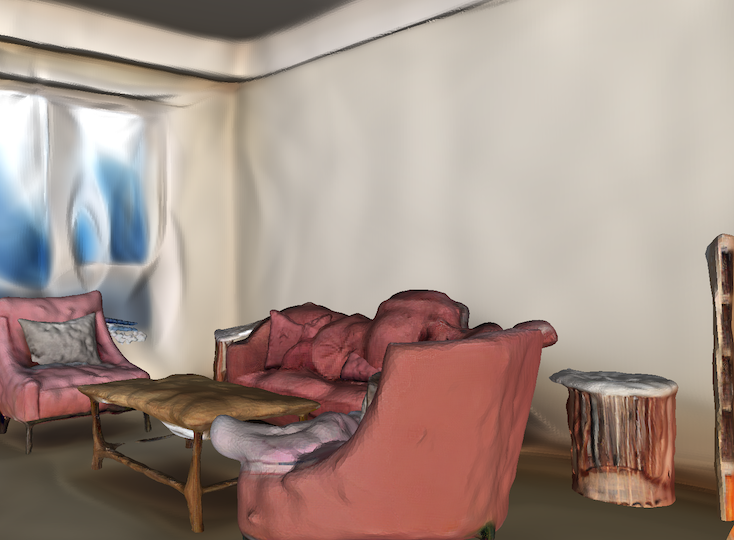} &
        \hspace{-3mm}\includegraphics[width=\wid]{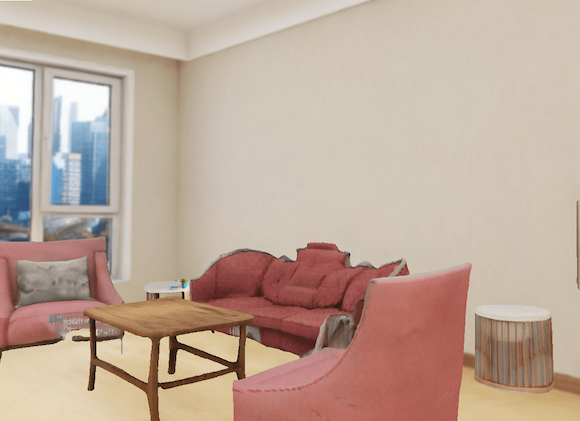} 
        \\ 
        \includegraphics[width=\wid]{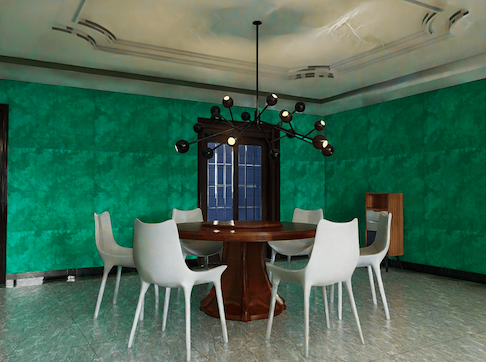} &
        \hspace{-3mm}\includegraphics[width=\wid]{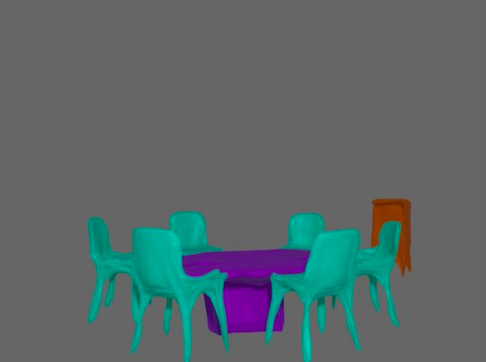}&
        \hspace{-3mm}\includegraphics[width=\wid]{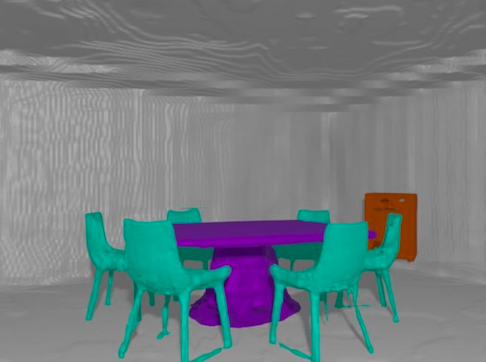} & 
        \hspace{-3mm}\includegraphics[width=\wid]{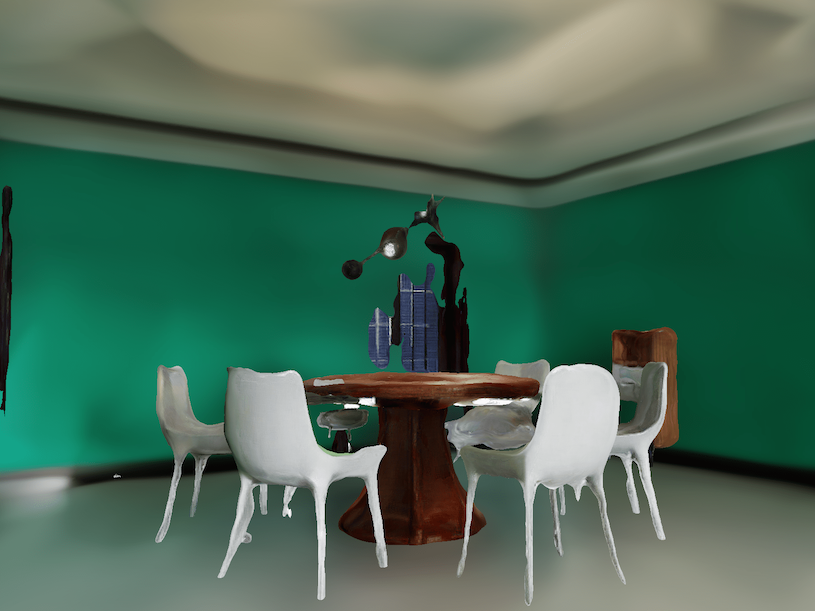} &
        \hspace{-3mm}\includegraphics[width=\wid]{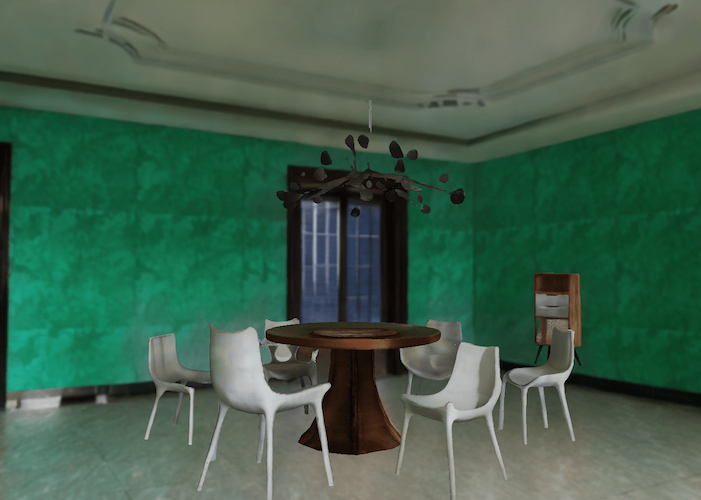} 
        \\ 
        \includegraphics[width=\wid]{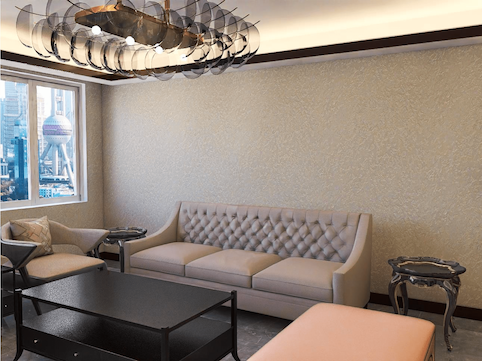} &
        \hspace{-3mm}\includegraphics[width=\wid]{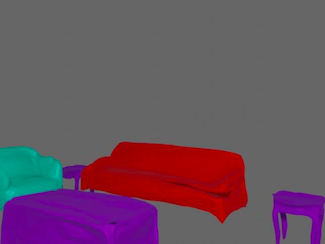}&
        \hspace{-3mm}\includegraphics[width=\wid]{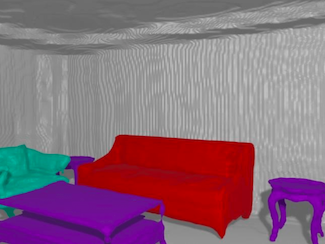} & 
        \hspace{-3mm}\includegraphics[width=\wid]
        {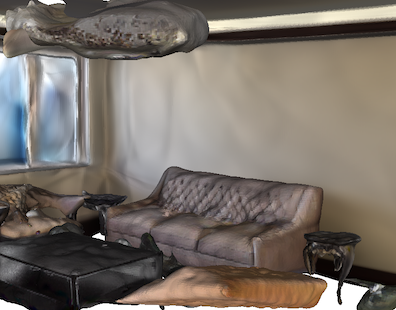} &
        \hspace{-3mm}\includegraphics[width=\wid]{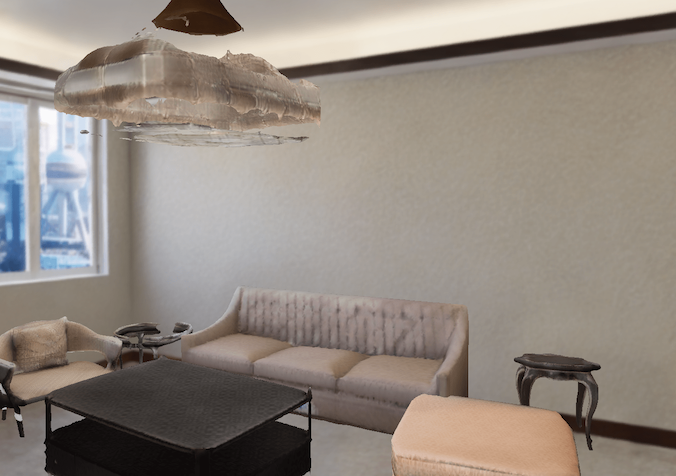} 
        \\ 
        \includegraphics[width=\wid]{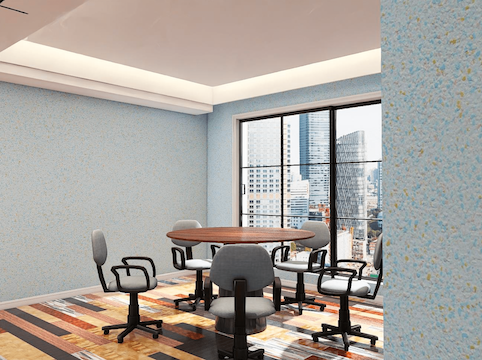} &
        \hspace{-3mm}\includegraphics[width=\wid]{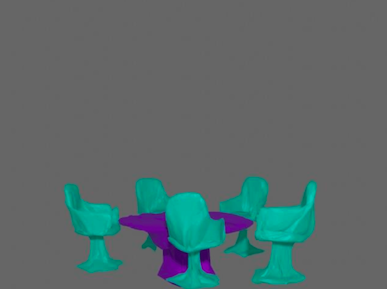}&
        \hspace{-3mm}\includegraphics[width=\wid]{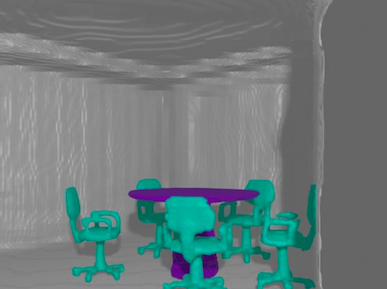} & 
        \hspace{-3mm}\includegraphics[width=\wid]
        {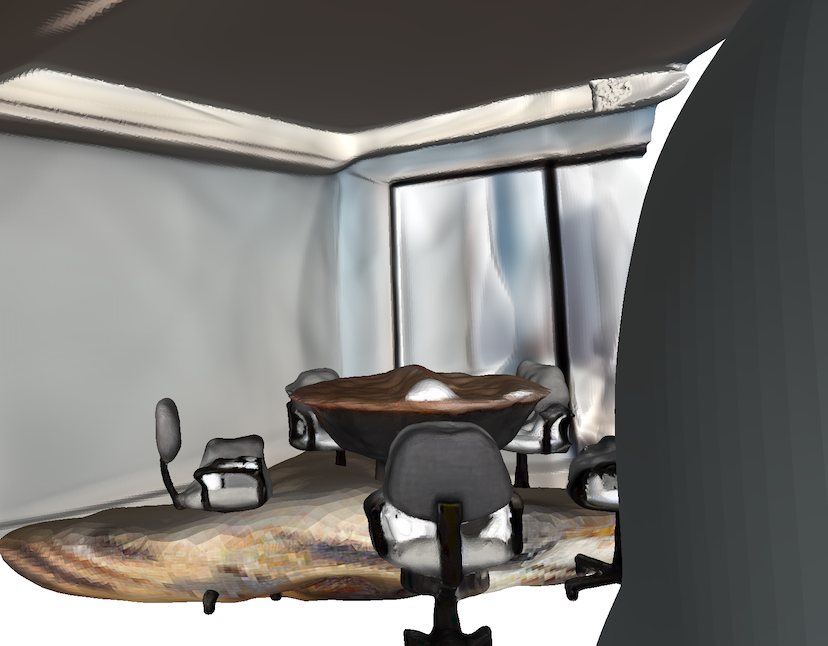} &
        \hspace{-3mm}\includegraphics[width=\wid]{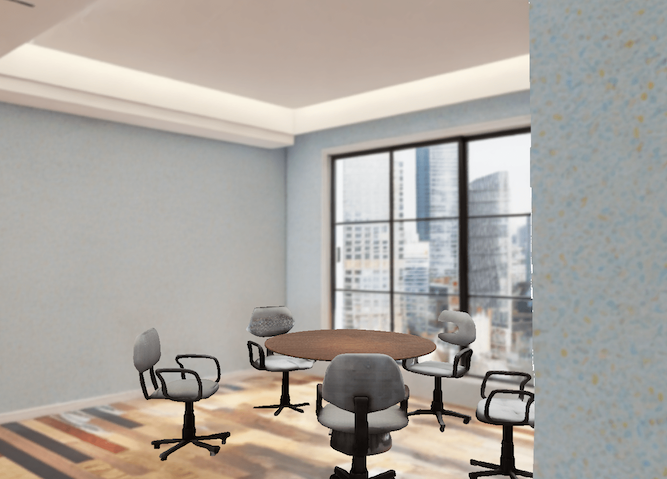}
        \\ 
        Input image & 
        Total3D~\cite{nie_total3dunderstanding_2020} & 
        InstPIFu~\cite{liu2022towards} & 
        Gen3DSR~\cite{dogaru_generalizable_2024} & 
        Ours
    \end{tabular}
    \caption[Qualitative Comparison of Holistic Scene Reconstruction on the 3D-FRONT dataset~\cite{fu20213d}]{\textbf{Qualitative Comparison of Holistic Scene Reconstruction} on the 3D-FRONT dataset~\cite{fu20213d}} 
    \label{fig:3dfront_qua2}
    \vspace{-0.3cm}
\end{figure*}

Table~\ref{tab:quantitative3dfont} demonstrates that our method achieves the best performance in terms of geometric accuracy compared to other methods in holistic scene reconstruction, largely thanks to our amodal completion module that contributes to the reconstruction of complete objects and the room background inpainting that correctly restores the occluded background region. This result is further validated by the visual performance of both datasets. On the 3D-Front dataset,~\cref{fig:3dfront_qua1},~\cref{fig:3dfront_qua2} show that InstPIFu~\cite{liu2022towards}, which reconstructs both foreground and background instances using an instance-aligned implicit function, achieving the highest overall mesh shape fidelity for foreground objects primarily due to its training on the 3D-Front dataset. However, their approach has notable limitations, including uneven mesh surfaces and a lack of texture in both foreground and background meshes, which restricts its applicability and may require additional post-processing to enhance visual realism. On the other hand, Gen3DSR~\cite{dogaru_generalizable_2024} improves the background reconstruction of InstPIFu by including colour information and reconstructs foreground instances with a similar approach to ours, which greatly enhances visual realism for the reconstructed scene. Nevertheless, the imprecise amodal completion module sometimes results in distorted shapes of foreground objects and persistent uneven surface issues (~\cref{fig:3dfront_qua2}), thus producing a less convincing and visually unpleasant scene reconstruction. Total3D~\cite{nie_total3dunderstanding_2020} produces textureless and non-watertight meshes that frequently lack detail, leading to the poorest visual results.

\begin{figure*}[htb]
    \centering    
    \setlength{\wid}{0.195\textwidth}
    \setlength{\mrg}{-0.2cm}
    \begin{tabular}{@{}ccccc@{}}
        \includegraphics[width=\wid]{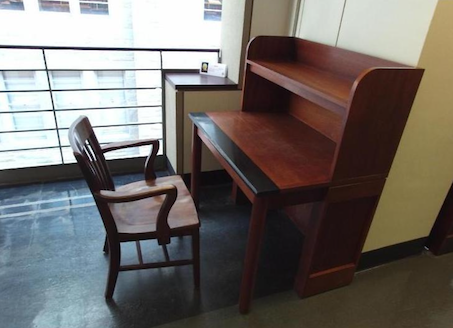} &
        \hspace{-3mm}\includegraphics[width=\wid]{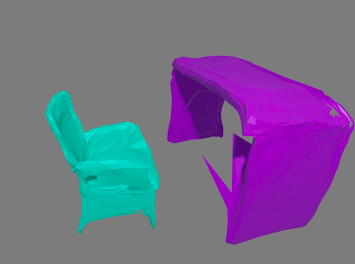}&
        \hspace{-3mm}\includegraphics[width=\wid]{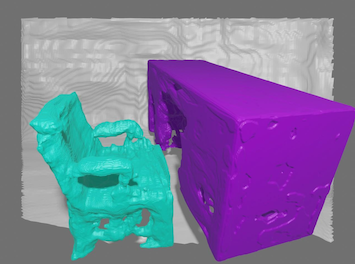} & 
        \hspace{-3mm}\includegraphics[width=\wid]
        {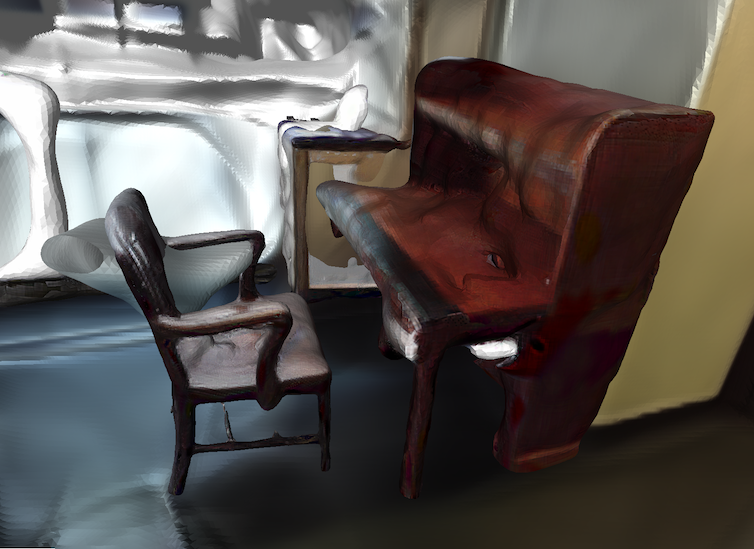} &
        \hspace{-3mm}\includegraphics[width=\wid]{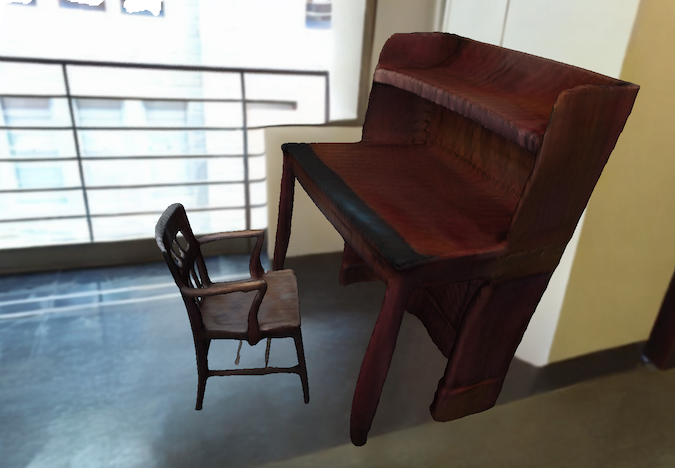} 
        \\ 
        \includegraphics[width=\wid]{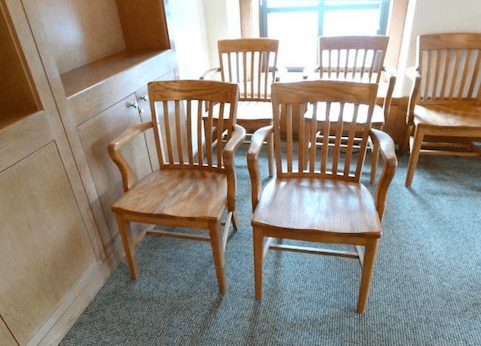} &
        \hspace{-3mm}\includegraphics[width=\wid]{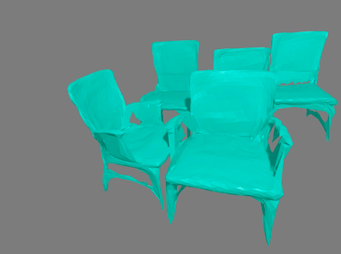}&
        \hspace{-3mm}\includegraphics[width=\wid]{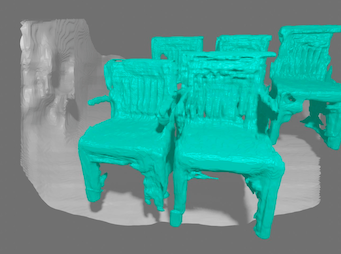} & 
        \hspace{-3mm}\includegraphics[width=\wid]
        {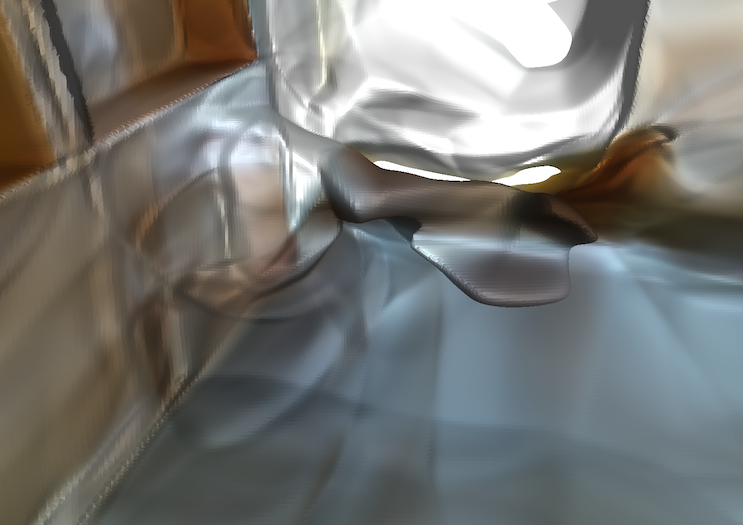} &
        \hspace{-3mm}\includegraphics[width=\wid]{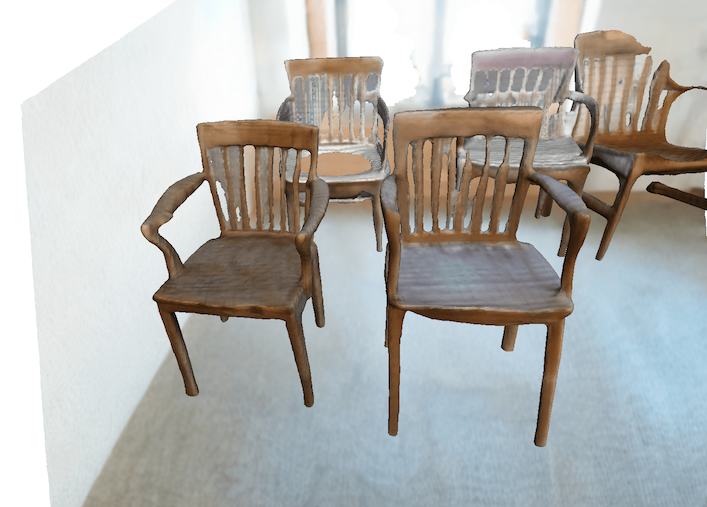} 
        \\ 
        \includegraphics[width=\wid]{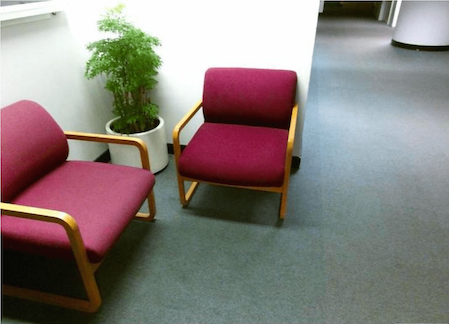} &
        \hspace{-3mm}\includegraphics[width=\wid]{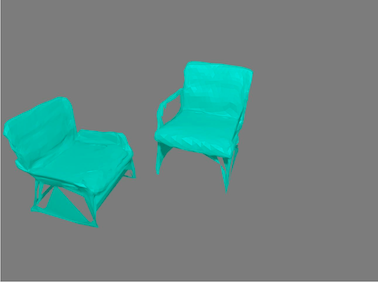}&
        \hspace{-3mm}\includegraphics[width=\wid]{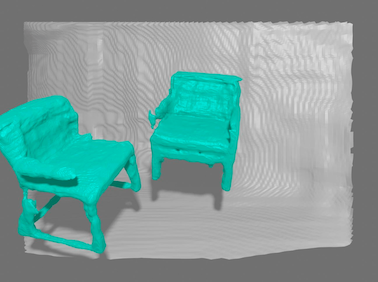} & 
        \hspace{-3mm}\includegraphics[width=\wid]
        {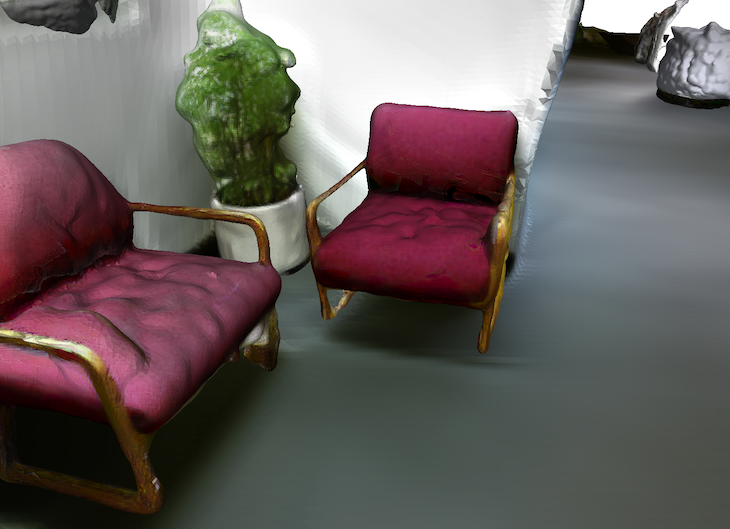} &
        \hspace{-3mm}\includegraphics[width=\wid]{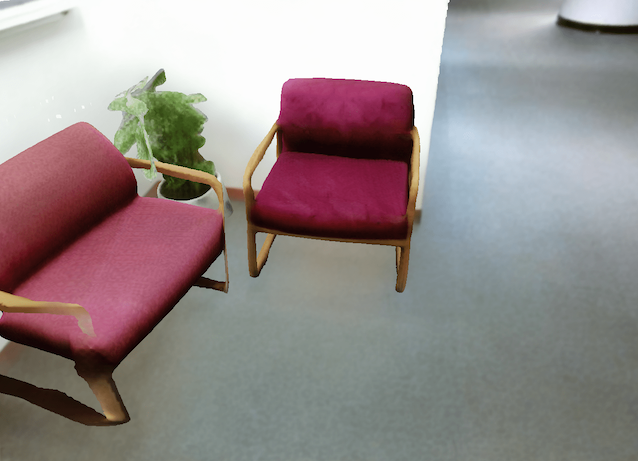}
        \\ 
        \includegraphics[width=\wid]{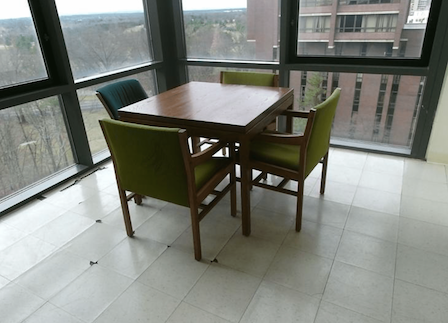} &
        \hspace{-3mm}\includegraphics[width=\wid]{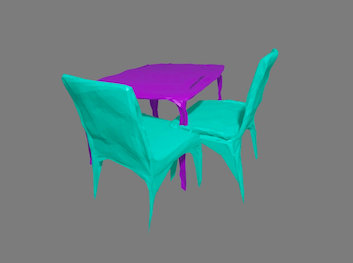}&
        \hspace{-3mm}\includegraphics[width=\wid]{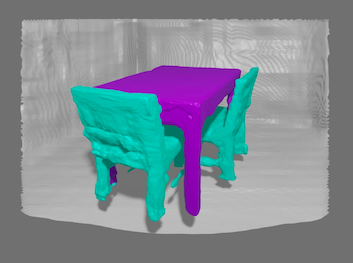} & 
        \hspace{-3mm}\includegraphics[width=\wid]
        {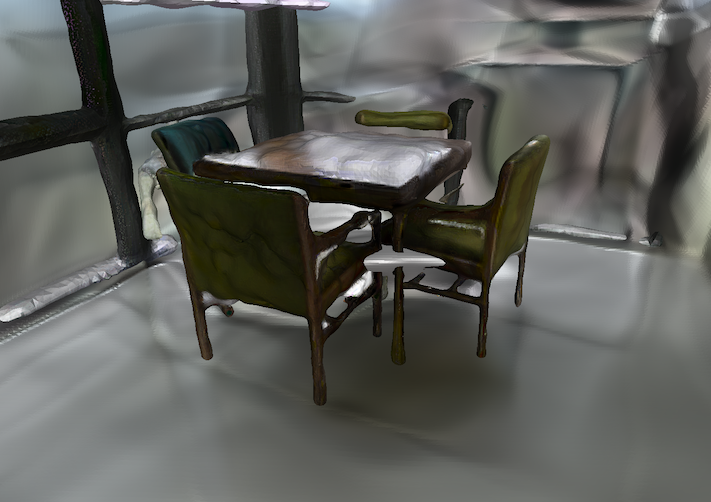} &
        \hspace{-3mm}\includegraphics[width=\wid]{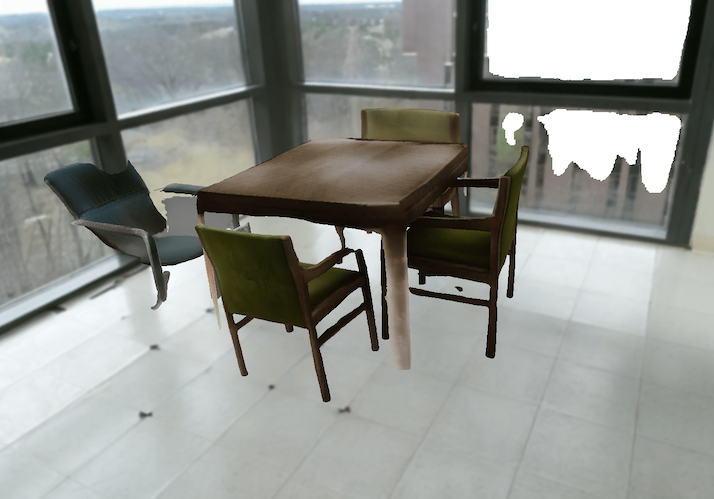}
        \\ 
        Input image & 
        Total3D~\cite{nie_total3dunderstanding_2020} & 
        InstPIFu~\cite{liu2022towards} & 
        Gen3DSR~\cite{dogaru_generalizable_2024} & 
        Ours
    \end{tabular}
    \caption[Qualitative Comparison of Holistic Scene Reconstruction on the SUN RGB-D dataset~\cite{zhou_learning_2014}]{\textbf{Qualitative Comparison of Holistic Scene Reconstruction} on the SUN RGB-D dataset~\cite{zhou_learning_2014}} 
    \label{fig:sunrgbd}
    \vspace{-0.3cm}
\end{figure*}

The results on the SUN RGB-D datasets (~\cref{fig:sunrgbd}) show that our method not only outperforms other learning-based methods, which often struggle with unseen furniture shapes and oversimplify them, but also generates higher quality foreground instances compared to Gen3DSR, due to our amodal completion module. In contrast, our method uses the rich prior embedded in stable-diffusion to complete the occluded view and handle diverse and novel furniture shapes and categories, which demonstrates its superior generalization ability.

\subsection{Ablation Study}
\begin{table}[b!]
\centering
\begin{tabular}{lcc}
\toprule
\textbf{Method} & \textbf{CD $\downarrow$} & \textbf{F-Score $\uparrow$} \\
\midrule
Ours w/o Marigold-Dust3r Fine-Tuning & 0.136 & 79.05 \\
Ours w/o Amodal Completion & 0.134 & 78.79 \\
Ours & \textbf{0.097} & \textbf{80.96} \\
\bottomrule
\end{tabular}
\caption[Ablation study of our Marigold depth fine-tuning and Amodal Completion modules]{\textbf{Ablation study} of our Marigold depth fine-tuning and Amodal Completion modules.}
\label{tab:ablation}
\end{table}
Our method follows a compositional strategy without direct training on instance-level 3D generation and 3D object detection. Post-processing steps, such as amodal completion for restoring the full view of furniture and Marigold-dust3r fine-tuning for better view-space alignment play a crucial role. To better understand the impact of these components, we conduct an ablation study to evaluate their contributions to the overall reconstruction performance. As shown in table~\ref{tab:ablation}, the inclusion of these two modules increases the overall quality of the reconstruction. The amodal completion module restores the complete view of foreground objects. Without amodal completion, the partial view of the instance is directly processed by image-to-3D models, potentially resulting in malformed 3D shapes. Fine-tuning Marigold with dust3r generates a depth map that translates into a 3D model with accurate geometry and fine details. In contrast, using the original Marigold depth with estimated scale and shift can lead to warped room geometry, which provides poorer guidance for view-space alignment than the fine-tuned version. 


\chapter{Discussion}

\section{Limitations}
While our method demonstrates good performance across various scenarios, certain limitations need to be addressed. A key challenge is its sensitivity to extreme occlusions. Although the amodal completion strategy successfully handles moderate occlusions, it can struggle when significant portions of an instance are obscured, leading to incomplete or less accurate reconstructions of foreground instances.

Another limitation arises in the depth estimation and image-to-3D generation processes, particularly when dealing with images captured using a field of view (FOV) that deviates significantly from the standard FOV used during the training of those models. Monocular depth estimation models are typically trained on images with a standard FOV, and when presented with wide-angle images, these models may produce less accurate depth maps due to the distortion inherent in such images. This distortion can result in warped or incorrect geometry in the reconstructed scene. Similarly, the image-to-3D generation process may as well be less accurate for images with non-standard FOVs, as the distortion affects the perception of instance shapes and sizes.

\section{Future Work}
Although our current framework is specialized for indoor scene reconstruction, there is potential to extend its applicability to a broader range of environments. Future work could explore adapting our method to outdoor scenes, urban environments, and other complex settings. These scenarios present unique challenges, such as a broader range of instance types and unbounded scenes, which would require further refinement of our semantic understanding and depth estimation modules.

Another promising direction is the integration of sparse view reconstruction. While our method is currently designed for single-view input, extending it to handle multiple viewpoints could greatly enhance the accuracy and robustness of the reconstruction process. With multiple views, we can better address issues related to scale, occlusion, and perspective distortion, providing a more reliable estimate of the true dimensions and spatial arrangement of instances within the scene.

\section{Potential Applications}
The proposed framework for single-view indoor scene 3D reconstruction has several potential applications across various fields:

\begin{itemize}
    \item \textbf{Cultural Heritage Preservation:} Museums and cultural institutions can use our framework to digitally reconstruct and preserve historical interiors from archival photographs. This application aids in the documentation and restoration of cultural heritage sites, which helps preserve valuable historical environments for future generations.
    
    \item \textbf{Interior Design:} Designers and home staging professionals can use this framework to create 3D models of rooms from a single photo, and easily edit them. This allows for rapid visualization of interior design changes or staging options without needing extensive 3D scanning equipment.

    \item \textbf{Real Estate Marketing:} Real estate agents can quickly generate 3D models of interiors from listing photos, providing potential buyers with an interactive and immersive way to explore properties online. This is particularly useful for showcasing properties to remote buyers or during the initial browsing stages.

\end{itemize}

\chapter{Conclusion}
In this thesis, we propose a novel modular framework for single-view indoor scene 3D reconstruction to address the limitations of traditional approaches that often struggle in complex environments with heavy occlusions and unseen instance shapes. By deconstructing the problem into distinct components—instance segmentation, depth estimation, amodal completion, inpainting for room layouts, image-to-3D, and view-space alignment—we have introduced a robust method that employs diffusion-based techniques to generate high-quality 3D reconstruction from a single 2D image.

Our framework holds potential for various applications, including interior design, real estate, and cultural heritage preservation, where single-view reconstruction can significantly enhance accessibility. However, despite its strengths, our framework does have limitations, particularly in handling extreme occlusions and non-standard field-of-view images. These challenges highlight areas for future research, where improvements in occlusion handling and adaptation to different environments, such as outdoor or urban settings, could further enhance the applicability of the proposed method. We believe our work establishes the foundation for more reliable and generalizable single-view 3D reconstruction, paving the way for its deployment in real-world applications and setting a new benchmark for future developments in this field.

\printbibliography
\end{document}